\documentclass[lettersize,journal]{IEEEtran}

\usepackage{times}
\usepackage{graphicx}
\usepackage{amssymb}
\usepackage{balance}
\usepackage{amsmath}
\usepackage{lipsum}  
\usepackage{adjustbox}
\usepackage{hyperref}
\usepackage{url}
\usepackage{lipsum}  
\usepackage{multicol}
\usepackage{multirow, makecell}
\usepackage{threeparttable}
\usepackage{booktabs}
\usepackage{pgfplots}
\usepackage{subcaption}
\usepgfplotslibrary{groupplots}
\usepackage{xcolor}
\pgfplotsset{width=\linewidth,compat=1.9}
\allowdisplaybreaks

\title{Grounded Relational Inference: Domain Knowledge Driven Explainable Autonomous Driving}

\author{Chen Tang$^{1,2\dagger}$, Nishan Srishankar$^{1\dagger}$, Sujitha Martin$^{1\dagger}$, Masayoshi Tomizuka$^{2}$
\thanks{$^{1}$ Honda Research Institute, CA, USA}
\thanks{$^{2}$ Department of Mechanical Engineering, UC Berkeley, CA, USA}
\thanks{$^{\dagger}$ Part of the work was conducted while Chen Tang and Nishan Srishankar were with Honda Research Institute as research interns. Chen Tang is currently with the Department of Computer Science at UT Austin, TX, USA. Nishan Srishankar is currently with JPMorgan Chase, NY, USA. Sujitha Martin is currently with Amazon Web Services, CA, USA.}

}

\begin{document}

\maketitle
\begin{abstract}
    Explainability is essential for autonomous vehicles and other robotics systems interacting with humans and other objects during operation. Humans need to understand and anticipate the actions taken by machines for trustful and safe cooperation. In this work, we aim to develop an explainable model that generates explanations consistent with both human domain knowledge and the model's inherent causal relation. In particular, we focus on an essential building block of autonomous driving\textemdash multi-agent interaction modeling. We propose Grounded Relational Inference (GRI). It models an interactive system's underlying dynamics by inferring an interaction graph representing the agents' relations. We ensure a semantically meaningful interaction graph by grounding the relational latent space into semantic interactive behaviors defined with expert domain knowledge. We demonstrate that it can model interactive traffic scenarios under both simulation and real-world settings and generate semantic graphs explaining the vehicle's behavior through their interactions.
    
\end{abstract}



\IEEEpeerreviewmaketitle


\section{Introduction}
\label{sec:introduction}
    \IEEEPARstart{D}{eep} learning has been utilized to address various autonomous driving problems~\cite{bojarski2016end,chen2017multi,chen2019dob}. However, deep neural networks lack the transparency that helps people understand their underlying mechanisms. It is a crucial drawback for safety-critical applications with humans involved (e.g., autonomous vehicles). Humans need to understand and anticipate the actions taken by machines for trustful and safe cooperation. In response to this problem, the concept of explainable AI (XAI) was introduced. It refers to machine learning techniques that provide details and reasons that make a model's mechanism easy to understand~\cite{arrieta2020explainable}. Most of the existing works for deep learning models focus on post-hoc explanations~\cite{arrieta2020explainable}. They enhance model explainability by unraveling the underlying mechanisms of a trained model: Vision-based approaches, such as visual attention~\cite{kim2017interpretable} and deconvolution~\cite{bojarski2018visualbackprop}, illustrate which segments of the input image affect the outputs; Interaction-aware models, such as social LSTM with social attention~\cite{alahi2016social,vemula2018social} and graph neural networks (GNN) with graph attention~\cite{hoshen2017vain, velivckovic2017graph, sukhbaatar2016learning, kipf2018neural}, identify the agents that are critical to the decision-making procedure.
        
    Although promising, post-hoc explanations could be ambiguous and falsely interpreted by humans. \textcolor{black}{For instance, a visual attention map only illustrates which regions of the input image the output of the model depends on. The semantic meaning behind the causal relation is left for human users to interpret. Kim et al.~\cite{kim2018textual} attempted to resolve the ambiguity by aligning textual explanations with visual attention. However, the underlying mechanism of the model is not necessarily consistent with the textual explanations. To truly build trust with humans, we argue that a deep learning model for an autonomous system should be equipped with explanations consistent with both \emph{human domain knowledge} and the model's \emph{inherent causal relation}}. 
    
    \textcolor{black}{In this work, we explore how to approach such an explainable model for an essential building block of autonomous driving\textemdash multi-agent interaction modeling. In particular, we focus on the relational inference problem studied in~\cite{kipf2018neural}. Kipf et al. propose the Neural Relational Inference (NRI) model, which models an interactive system by explicitly inferring its inherent interactions. Formally, the NRI model aims to solve a reconstruction task. Given the observed trajectories of all the objects, an encoder first infers the interactions between objects represented by a latent interaction graph, whose edges are aligned with discrete latent variables corresponding to a cluster of pairwise interaction behaviors between the objects. Afterward, a decoder learns the dynamical model conditioned on the inferred interaction graph and then reconstructs the trajectories given the initial states. If the decoder can accurately reconstruct the trajectories, it indicates that the latent space effectively models the interactions.}
    
    \textcolor{black}{We find this discrete latent space interesting because the inferred interaction graph could potentially serve as an explanation directly: it explains the reconstructed trajectories as a sequence of interaction behaviors among agents. Moreover, the reconstructed trajectories are governed by the same interaction graph. Therefore, the NRI model seems promising to fulfill our goal to make the explanation consistent with the model's underlying mechanism. However, since the NRI model learns the latent space in an unsupervised manner, it is difficult for humans to interpret the semantic meaning behind those interaction behaviors, which makes the interaction graph ambiguous as an explanation. To address this issue, we propose to ground the latent space in a set of interactive behaviors defined with human domain knowledge.}
     
    \begin{figure*}[t]
	\centering
	\includegraphics[width=5in]{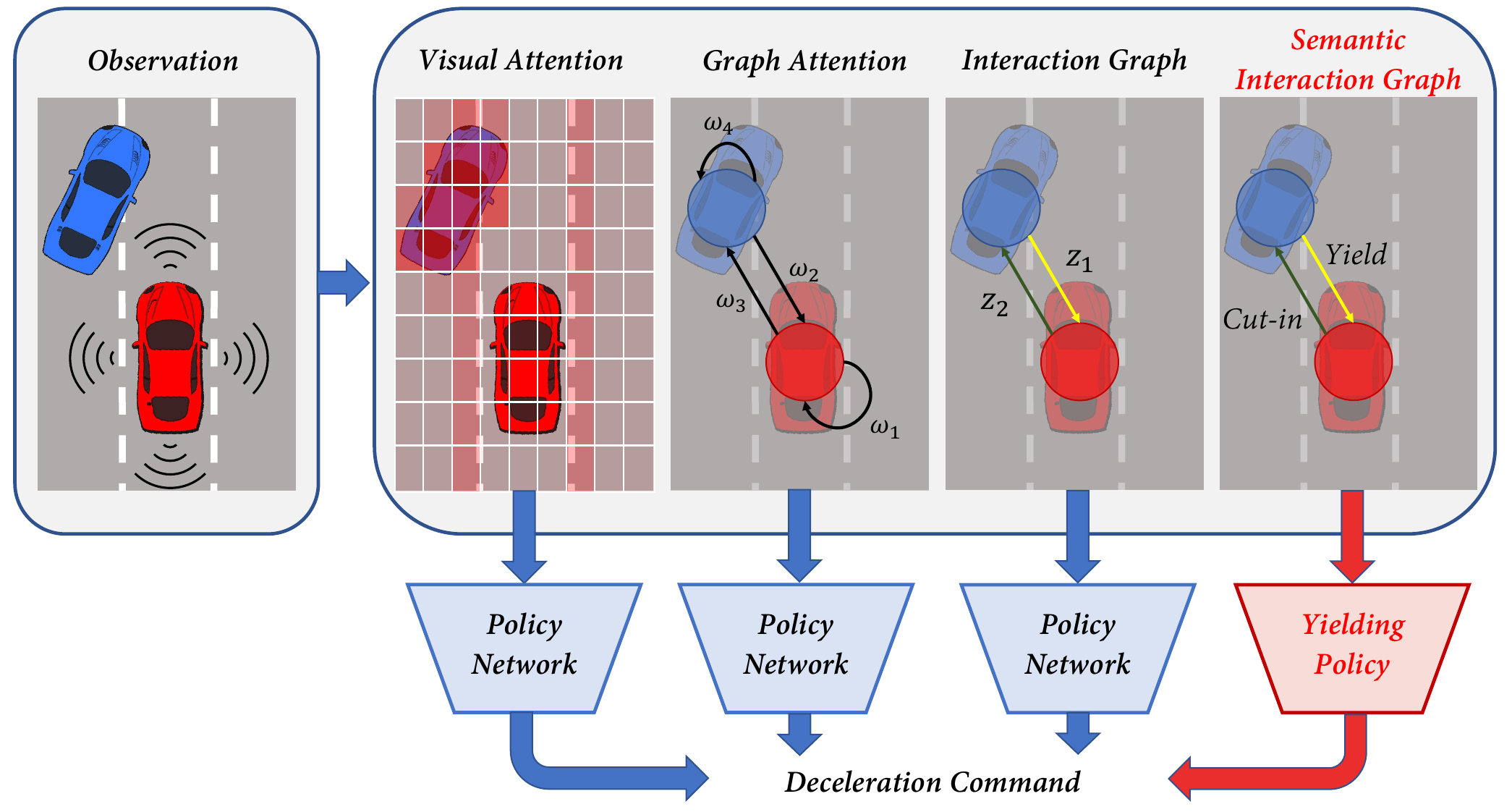}
	\caption{A motivating lane-changing scenario where we ask different models to control the red vehicle. All the models generate deceleration commands but have different intermediate outputs. With the aid of visual attention, we generate a heat map indicating the critical pixels of the input image. Graph attention network assigns edge weights $\omega_i$ to specify the importance of surrounding vehicles to the controlled vehicle. However, the attention mechanisms cannot recognize different effects\textemdash the two cars are mutually important but affect each other in distinct ways. The NRI model can distinguish between different interactive behaviors by assigning different values to the latent variables $z_i$ in the interaction graph. Still, the latent space does not have explicit semantic meaning. In contrast, our model ensures a semantic interaction graph, which illustrates the model's understanding of the scenario and explains the action it takes. It determines the interaction graph with a latent space grounded in yielding and cutting-in behaviors. It learns the control policies that generate behaviors consistent with their definitions in domain knowledge (e.g., traffic rules) and executes the corresponding policies according to the inferred edge types.}
	\label{fig:example}
    \end{figure*}
    
    As a running example, consider the scenario depicted in Fig.~\ref{fig:example}, where we ask different models to control the red vehicle. Attention mechanisms can indicate the critical pixels or agents, but they cannot recognize different effects\textemdash the two cars are mutually important but affect each other in distinct ways. The NRI model can distinguish between different interactive behaviors. Still, the latent space does not have explicit semantic meaning. In contrast, our model should determine the interaction graph with a latent space grounded in yielding and cutting-in behaviors. It learns the control policies that generate behaviors consistent with their definitions in domain knowledge (e.g., traffic rules) and executes the corresponding policies according to the inferred edge types. This semantic interaction graph illustrates the model's understanding of the scenario and explains the action it takes.
        
    \textcolor{black}{If we merely want to make the interaction graph consistent with humans' labeling of the scenes, a straightforward approach is training the encoder directly via supervised learning.} Interaction labels can be either obtained from human experts~\cite{sun2018probabilistic} or rule-based labeling functions~\cite{lee2019joint}. \textcolor{black}{However, labels for the interaction graph are insufficient to induce the decoder to synthesize the interactive behaviors suggested by the labels, because the model cannot capture the semantic meaning behind those interaction labels.} Instead, we recast relational inference into an inverse reinforcement learning (IRL) problem and introduce structured reward functions to ground the latent space. Concretely, we model the system as a multi-agent Markov decision process (MDP), where the agents share a reward function that depends on the relational latent space. We design structured reward functions based on expert domain knowledge to explicitly define the interactive behaviors corresponding to the latent space. To solve the formulated IRL problem, we propose Grounded Relational Inference (GRI). It has a variational-autoencoder-like (VAE) GNN in NRI~\cite{kipf2018neural} as the backbone model. Additionally, we incorporate the structured reward functions into the model as an additional reward decoder. A variational extension of the adversarial inverse reinforcement learning (AIRL) algorithm is derived to train all the modules simultaneously.
    
    \textcolor{black}{Compared to direct supervision via interaction labels, we provide implicit supervision to GRI in terms of the structures of the reward functions. Since each reward function defines a type of interactive behavior, we confine the latent space to a cluster of interactive behaviors. It mainly has two advantages over supervision through labeling: 1) First, since the policy decoder learns to maximize the cumulative reward given the inferred interaction graph, the structured reward functions guide the policy to synthesize the corresponding semantic behaviors rather than simply mimicking the demonstrated trajectories; 2) Second, the end-to-end training scheme leaves the model to identify the underlying interaction graph of the observed trajectories and learn the characteristics of different behaviors (i.e., parameters of reward functions) from data. It avoids the undesired bias introduced during the labeling procedure. Labels generated by human experts are subjective. Different people may interpret an interacting scenario in different ways. In contrast, there exist systematic and principled ways to investigate what reward functions human behavior is subject to from data~\cite{naumann2020analyzing}.}
    
    \textcolor{black}{The remaining content is organized as follows. In Section~\ref{sec:related-work}, we give a concise review of existing works that are closely related to ours in terms of methodology or motivation. In Section~\ref{sec:background}, we briefly summarize NRI and AIRL to prepare the readers for the core technical content. In Section~\ref{sec:formulation}, we introduce how we reformulate relational inference into a multi-agent IRL problem with relational latent space. In Sec~\ref{sec:method}, we present the GRI model in a general context. In Section~\ref{sec:experiments}, we demonstrate how we apply the proposed framework to model some simple traffic scenarios in both simulation and real-world settings. The experimental results show that GRI can model interactive traffic scenarios, and generate semantic interaction graphs that are consistent with both human domain knowledge and the modeled interactive behaviors.}
    
    \section{Related Work}
    \label{sec:related-work}
    Our model combines graph neural networks and adversarial inverse reinforcement learning for interactive system modeling. This section gives a concise review of these two topics and summarizes the existing works closely related to ours. We also discuss some additional works on explainable driving models as a complement to the discussion in Sec.~\ref{sec:introduction}.
        
    {\bf Interaction modeling using GNN.} \textcolor{black}{GNN refers to the class of neural networks that operate on graph-structured data, which consist of local message-passing functions shared across all nodes to aggregate the information of neighboring nodes. The inductive biases introduced by the shared message-passing operations make GNN particularly effective for modeling multi-agent interactive systems~\cite{sukhbaatar2016learning, van2018relational, battaglia2016interaction}. Multi-agent systems can be naturally represented as graphs. Furthermore, sharing the message-passing functions across nodes improves learning efficiency and enables modeling systems with varying numbers of agents with a single GNN.} One category of models we find interesting is those with graph attention mechanisms. One seminal work is Graph Attention Network (GAT)~\cite{velivckovic2017graph}, which performed well on large-scale inductive classification problems. VAIN~\cite{hoshen2017vain} applied attention in multi-agent modeling. The attention map unravels the interior interaction structure to some extent which improves the explainability of VAIN. An approach closely related to ours is NRI~\cite{kipf2018neural}, which modeled the interaction structure explicitly with discrete relational latent space compared to the continuous graph attention. We explain the difference between NRI and our proposed method in Sec.~\ref{sec:introduction} and~\ref{sec:method}. One related work in the autonomous driving domain is~\cite{lee2019joint}, which also modeled interactive driving behavior with semantically meaningful interactions but in a supervised manner. 
    
    \textcolor{black}{Another type of model we want to mention is the spatio-temporal graphs (st-graph). St-graph decomposes a complex problem into components and their spatio-temporal interactions, which are represented by nodes and edges of a factor graph. It makes st-graph a ubiquitous representation for interacting systems, e.g., human motion~\cite{jain2016structural}, human-robot interaction~\cite{liu2021decentralized}, and traffic flow~\cite{yu2017spatio}. Jain et al.~\cite{jain2016structural} proposed a general method to transform any st-graph to a mixture of RNNs called structural-RNN (S-RNN). When using GRUs, our GNN policy decoder is similar to S-RNN, as they capture the same spatio-temporal dependency. In particular, Liu et al.~\cite{liu2021decentralized} combined S-RNN with model-free RL to obtain a structured policy for robot crowd navigation. In terms of the underlying MDP, our GRI model is developed based on a multi-agent MDP, whereas theirs has a single robot as the agent and regards the surrounding humans as parts of the environment. In addition, we adopt a structured reward function for each agent based on the graph and introduce a relational latent space into the MDP. }
    
    \textcolor{black}{{\bf Adversarial IRL and Imitation Learning.} Now we give a brief review of related works on adversarial IRL. We also include prior works related to generative adversarial imitation learning (GAIL)~\cite{ho2016generative}, because GAIL is closely connected to AIRL~\cite{finn2016connection}. Both methods have GANs as the backbone models, and learn the discriminator through MaxEntIRL. The difference is that GAIL uses an unstructured discriminator and does not use the generator’s density.}
    
    Our work is mainly related to two categories of methods: multi-agent and latent AIRL/GAIL algorithms. Yu et al.~\cite{yu2019multi} proposed a multi-agent AIRL framework for Markov games under correlated equilibrium. It is capable of modeling general heterogeneous multi-agent interactions. The PS-GAIL algorithm~\cite{bhattacharyya2018multi} considered a multi-agent environment in the driving domain that is similar to ours\textemdash cooperative and homogeneous agents with shared policies\textemdash and extended GAIL~\cite{ho2016generative} to model the interactive behaviors. In~\cite{bhattacharyya2019simulating}, they augmented the reward in PS-GAIL as a principle manner to specify prior knowledge, which shares the same spirit with the structured reward functions in GRI. 
    
    Latent AIRL models integrate a VAE into either the discriminator or the generator for different purposes. Wang et al.~\cite{wang2017robust} conditioned the discriminator on the embeddings generated by a VAE trained separately using behavior cloning. The VAE encodes trajectories into low-dimensional space, enabling the generator to produce diverse behaviors from the limited demonstration. VDB~\cite{peng2018variational} constrained information contained in the discriminator's internal representation to balance the training procedure for adversarial learning algorithms. The PEMIRL framework~\cite{yu2019meta} achieved meta-IRL by encoding demonstration into a contextual latent space. Though studied in a different context, PEMIRL is conceptually similar to our framework as both its generator and discriminator depend on the inferred latent variables. 
    
    {\bf Explainable Autonomous Driving.} At the end of this section, we discuss some additional works related to explainable autonomous driving as a complement to those we have mentioned in Sec.~\ref{sec:introduction}. They addressed some shortcomings of the discussed approaches, especially those methods based on attention mechanisms. Kim et al.~\cite{kim2018textual} trained a textual explanation generator concurrently with a visual-attention-based controller in a supervised manner. It generates sentences explaining the control action as a consequence of certain objects highlighted in the attention map, which can be easily interpreted compared to visual attention. Another issue of attention that has been raised in the literature is causal confusion~\cite{de2019causal}. The model does not necessarily assign high attention weights to objects/regions that influence the control actions. In~\cite{kim2017interpretable}, a fine-grained decoder was proposed to refine visual attention maps and detect critical regions through causality tests. In~\cite{li2020make}, Li et al. adopted a similar idea for object-level reasoning. Causal inference was applied to identify risk objects in driving scenes. One interesting observation was that the detection accuracy was improved with intervention during the training stage, i.e., augmenting the training data by masking out non-causal objects. However, intervention requires explicit prior knowledge of the causal relations to label the casual and non-causal objects in a scene. Similar to intention labels, such kinds of labels are generally prohibitive to obtain due to the intricate nature of human cognition.  


\section{Background}
\label{sec:background}
In this section, we would like to briefly summarize two algorithms that are closely related to our approach, in order to prepare the readers for the core technical content.

    \subsection{Neural Relational Inference (NRI)}\label{subsec:nri}
    Kipf et al.~\cite{kipf2018neural} represent an interacting system with $N$ objects as a complete bi-directed graph $\mathcal{G_{\mathrm{scene}} = (V, E)}$ with vertices $\mathcal{V}=\left\{v_i\right\}_{i=1}^{N}$ and edges $\mathcal{E}=\left\{e_{i,j}=(v_i, v_j) \mid i \neq j \right\}$. The edge $e_{i,j}$ refers to the one pointing from the vertex $v_i$ to $v_j$. Each vertex corresponds to an object in the system. The NRI model is formalized as a VAE with a GNN encoder inferring the underlying interactions and a GNN decoder synthesizing the system dynamics given the interactions. 
    
    Formally, the model aims to reconstruct a given state trajectory, denoted by  $\mathbf{x}=\left(\mathbf{x}^0,\dots, \mathbf{x}^{T-1}\right)$, where $T$ is the number of timesteps and  $\mathbf{x}^t=\left\{\mathbf{x}^t_1,\dots,\mathbf{x}^t_N\right\}$. The vector $\mathbf{x}^t_i\in{\mathbb{R}^n}$ denotes the state vector of object $v_i$ at time $t$. Alternatively, the trajectory can be decomposed into $\mathbf{x}=(\mathbf{x}_1, \dots, \mathbf{x}_N)$, where $\mathbf{x}_i=\left\{\mathbf{x}^0_i,\dots,\mathbf{x}^{T-1}_i\right\}$. The encoder operates over $\mathcal{G}_\mathrm{scene}$, with $\mathbf{x}_i$ as the node feature of $v_i$. It infers the posterior distribution of the edge type ${z}_{i,j}$ for all the edges, collected into a single vector $\mathbf{z}$. The decoder operates over an interaction graph $\mathcal{G}_\mathrm{interact}$ and reconstructs $\mathbf{x}$. The graph $\mathcal{G}_\mathrm{interact}$ is constructed by assigning sampled $\mathbf{z}$ to the edges of $\mathcal{G}_\mathrm{scene}$ and assigning the initial state to the nodes of $\mathcal{G}_\mathrm{scene}$. If $\mathcal{G}_\mathrm{interact}$ represents the interactions sufficiently, the decoder should be able to reconstruct the trajectory accurately. 
    
    The model is trained by maximizing the evidence lower bound (ELBO):
    \begin{equation*}
        \mathcal{L}=\mathbb{E}_{q_\phi(\mathbf{z}\vert\mathbf{x})}\left[\log p_\gamma (\mathbf{x}\vert\mathbf{z})\right]-D_{KL} \left[q_\phi(\mathbf{z}\vert\mathbf{x})\vert\vert p (\mathbf{z})\right],
    \end{equation*}
    where $q_\phi(\mathbf{z}\vert\mathbf{x})$ is the encoder output which can be factorized as:
    \begin{equation}
        q_\phi(\mathbf{z}\vert\mathbf{x})=\prod_{i=1}^N\prod_{j=1, j\neq i}^N q_\phi(z_{i,j}\vert \mathbf{x}), \label{eqn:facto}
    \end{equation}
    where $\phi$ refers to the parameters of the encoder. The decoder output $p_\gamma(\mathbf{x}\vert\mathbf{z})$ can be written as:
    \begin{equation*}
        p_\gamma(\mathbf{x}\vert\mathbf{z})=\prod_{t=0}^{T-1}p_\gamma(\mathbf{x}^{t+1}\vert{\mathbf{x}^t, \dots, \mathbf{x}^0, \mathbf{z}}),
    \end{equation*}
    where $\gamma$ refers to the parameters of the decoder. 
    
    \subsection{Adversarial Inverse Reinforcement Learning (AIRL)}\label{subsec:airl}
    The AIRL algorithm follows the principle of maximum entropy IRL~\cite{ziebart2008maximum}. Consider an MDP defined by $(\mathcal{X, A, T}, r)$, where $\mathcal{X, A}$ are the state space and action space respectively. In the rest of the paper, we use $\mathbf{x}$ and $\mathbf{a}$ with any superscript or subscript to represent a state and action in $\mathcal{X}$ and $\mathcal{A}$. $\mathcal{T}$ is the transition operator given by $\mathbf{x}_{t+1}=f(\mathbf{a}_t, \mathbf{x}_t)$\footnote{The transition is assumed deterministic to simplify the notation. A more general form of the algorithm can be derived for stochastic systems, which is essentially the same with the deterministic case.}, and $r:\mathcal{X} \times \mathcal{A}\rightarrow \mathbb{R}$ is the reward function. The maximum entropy IRL framework assumes a suboptimal expert policy $\pi^\mathrm{E}(\mathbf{a}\vert\mathbf{x})$. The demonstration trajectories generated with the expert policy, $\mathcal{D^\mathrm{E}}=\left\{\boldsymbol{\tau}^\mathrm{E}_1, \dots \boldsymbol{\tau}^\mathrm{E}_M\right\}$ where $\boldsymbol{\tau}^\mathrm{E}_{i}=\left(\mathbf{x}_i^{\mathrm{E}, 0},\mathbf{a}_i^{\mathrm{E}, 0}, \dots, \mathbf{x}_i^{\mathrm{E}, T-1}, \mathbf{a}_i^{\mathrm{E}, T-1}\right)$, have probabilities increasing exponentially with the cumulative reward. Concretely, they follow a Boltzmann distribution: 
    \begin{equation*}
       \boldsymbol{\tau}^\mathrm{E}_i\sim{\pi^\mathrm{E}(\boldsymbol{\tau})} = \frac{1}{Z}\exp\left(\sum_{t=0}^{T-1} r_\lambda(\mathbf{x}_t, \mathbf{a}_t)\right),
    \end{equation*}
    where $r_\lambda$ is the reward function with parameters denoted by $\lambda$. Maximum entropy IRL aims to infer the underlying reward function parameters of the expert policy. It is formalized as a maximum likelihood problem:
    \begin{equation*}
        \lambda^* = \mathrm{arg} \max_\lambda \mathbb{E}_{\boldsymbol{\tau}^\mathrm{E}\sim\pi^\mathrm{E}(\boldsymbol{\tau})}\left[\sum_{t=0}^{T-1} r_\lambda(\mathbf{x}^\mathrm{E}_t, \mathbf{a}^\mathrm{E}_t)\right] - \log Z.
    \end{equation*}

    To derive a feasible algorithm to solve the problem, we need to estimate the partition function $Z$. One practical solution is co-training a policy model with the currently estimated reward function through reinforcement learning~\cite{finn2016guided}. Finn et al.~\cite{finn2016connection} found the equivalency between it and a special form of the generative adversarial network (GAN). The policy model is the generator, whereas a structured discriminator is defined with the reward function to distinguish a generated trajectory $\boldsymbol{\tau}^\mathrm{G}$ from a demonstrated one $\boldsymbol{\tau}^\mathrm{E}$. Fu et al.~\cite{fu2017learning} proposed the AIRL algorithm based on it, using a discriminator that identifies generated samples based on the pairs of state and action instead of the entire trajectory to reduce variance:
    \begin{equation}
        \mathcal{D}_{\lambda,\eta}(\mathbf{x},\mathbf{a})=\frac{\exp\left\{r_\lambda(\mathbf{x},\mathbf{a})\right\}}{\exp\left\{r_\lambda(\mathbf{x},\mathbf{a})\right\}+\pi_\eta(\mathbf{a}\vert\mathbf{x})}, \label{eqn:dis}
    \end{equation}
    where $\pi_\eta(\mathbf{a}|\mathbf{x})$ is the policy model with parameters denoted by $\eta$. The models $\mathcal{D}_{\lambda,\eta}$ and $\pi_\eta$ are trained adversarially by solving the following min-max optimization problem:  
    \begin{equation}
    \begin{split}
        \min_\eta \max_{\lambda} \quad & \mathbb{E}_{\mathbf{x}^\mathrm{E}, \mathbf{a}^\mathrm{E}\sim\pi^\mathrm{E}(\mathbf{x,a})}\left[\log\left(\mathcal{D}_{\lambda,\eta}(\mathbf{x}^\mathrm{E},\mathbf{a}^\mathrm{E})\right)\right] \\ 
        + & \mathbb{E}_{\mathbf{x}^\mathrm{G}, \mathbf{a}^\mathrm{G}\sim\pi_\eta(\mathbf{x,a})}\left[\log\left(1-\mathcal{D}_{\lambda,\eta}(\mathbf{x}^\mathrm{G},\mathbf{a}^\mathrm{G})\right)\right], \label{eqn:opt}   
    \end{split}
    \end{equation}
    where $\pi^\mathrm{E}(\mathbf{x,a})$ denotes the distribution of state and action induced by the expert policy, and $\pi_\eta(\mathbf{x,a})$ is the distribution induced by the learned policy.
    

\section{Problem Formulation}
\label{sec:formulation}
    Our GRI model grounds the relational latent space in a clustering of semantically meaningful interactions by reformulating the relational inference problem into a multi-agent IRL problem. Since the framework has the potential to be generalized to interactive systems in other domains apart from autonomous driving, we will introduce our approach in a general tone. However, it should be noted that we limit our discussion in this paper to autonomous driving problems, without claiming that it can be directly applied to other domains. GRI relies on expert domain knowledge to identify all possible semantic behaviors and design the corresponding reward functions. There exists a broad range of literature on interactive driving behavior modeling~\cite{sun2018probabilistic, kesting2010enhanced}, which we can refer to when designing the rewards. We can extend the proposed framework to other fields if proper domain knowledge is available, which is left for future investigation.
    
    We start with modeling the interactive system as a multi-agent MDP with graph representation. As in NRI, the system has an underlying interaction graph $\mathcal{G}_\mathrm{interact}$. The discrete latent variable $z_{i,j}$ takes a value from ${0, 1, \dots, K-1}$, where $K$ is the number of interactions. It indicates the type of relation between $v_i$ and $v_j$ with respect to its effect on $v_j$. Additionally, we assume the objects of the system are homogeneous intelligent agents who make decisions based on their interactions with others.
    
    Concretely, each of them is modeled with identical state space $\mathcal{X}$, action space $\mathcal{A}$, transition operator $\mathcal{T}$ and reward function $r:\mathcal{X} \times \mathcal{A}\rightarrow \mathbb{R}$. At time step $t$, the reward of agent $v_j$ depends on the states and actions of itself and the pairwise interactions between itself and all its neighbors:
    \begin{equation}
    \begin{split}
        &r_{\xi, \psi}(v^t_j, \mathbf{z}_j) = r_\xi^{n}(\mathbf{x}^t_j, \mathbf{a}^t_j) \\
        & \quad\quad\quad + \sum_{i\in\mathcal{N}_j}\sum_{k=1}^{K}\mathbf{1}(z_{i,j}=k) r^{{e},k}_{\psi_k}(\mathbf{x}^t_i, \mathbf{a}^t_i, \mathbf{x}^t_j, \mathbf{a}^t_j), \label{eqn:reward}    
    \end{split}
    \end{equation}
    where $\mathbf{z}_j$ is the collection of  $\left\{{z}_{i,j}\right\}_{i\in\mathcal{N}_j}$, $r_\xi^{n}$ is the node reward function parameterized by $\xi$, $\mathcal{N}_j$ is the set of $v_j$'s neighbouring nodes, $\mathbf{1}$ is the indicator function, and $r_{\psi_k}^{{e}, k}$ is the edge reward function parameterized by $\psi_k$ for the $k^\mathrm{th}$ type of interaction. We utilize expert domain knowledge to design $r_{\psi_k}^{{e}, k}$, so that the corresponding interactive behavior emerges by maximizing the rewards. Particularly, the edge reward equals zero for $k=0$, indicating that the action taken by $v_j$ does not depend on its interaction with $v_i$. 
    
    We assume the agents act cooperatively to maximize the cumulative reward of the system:
    \begin{equation*}
        \mathcal{R}_{\xi, \psi}(\boldsymbol{\tau},\mathbf{z})=\sum_{t=0}^{T-1}\mathbf{r}_{\xi, \psi}\left(\mathbf{x}^t, \mathbf{a}^t, \mathbf{z}\right)
        =\sum_{t=0}^{T-1}\sum_{j=1}^{N}r_{\xi, \psi}\left(v^t_j, \mathbf{z}_j\right),
    \end{equation*}
    with a joint policy denoted by $\boldsymbol{\pi}_\eta \left(\mathbf{a}^t\vert{\mathbf{x}^t, \mathbf{z}}\right)$. \textcolor{black}{The cooperative assumption does not generally hold for real-world traffic scenarios. Each individual driver may value the utilities differently. In practice, though, it has been shown that realistic highway driving behavior can be learned from real-world data under the cooperative assumption~\cite{8793750, bhattacharyya2022modeling}. It implies that the assumption can approximately apply in practical circumstances, especially in the highway scenarios tested in this work. Thus, we adopt the cooperative assumption to simplify the problem formulation and training procedure. In future work, we will investigate a non-cooperative extension of the proposed method to tackle interactive traffic scenarios where the cooperative assumption has to be removed~\cite{yu2019multi}.}
    
    Given a demonstration dataset, we aim to infer the underlying reward function and policy. Different from a typical IRL problem, both $r_{\xi, \psi}$ and $\pi_{\eta}$ depend on $\mathbf{z}$. Therefore, we need to infer the distribution $p(\mathbf{z}\vert\boldsymbol{\tau})$ to solve the IRL problem.

    \section{Grounded Relational Inference}\label{sec:method}
    We now present the Grounded Relational Inference model to solve the IRL problem specified in Sec.~\ref{sec:formulation}. The model consists of three modules modeled by message-passing GNNs~\cite{gilmer2017neural}: an encoder inferring the posterior distribution of edge types, a policy decoder generating control actions conditioned on the edge variables sampled from the posterior distribution, and a reward decoder modeling the rewards conditioned on the inferred edge types.
    
    \subsection{Architecture}
\begin{figure*}[t]
	\centering
	\includegraphics[width=5in]{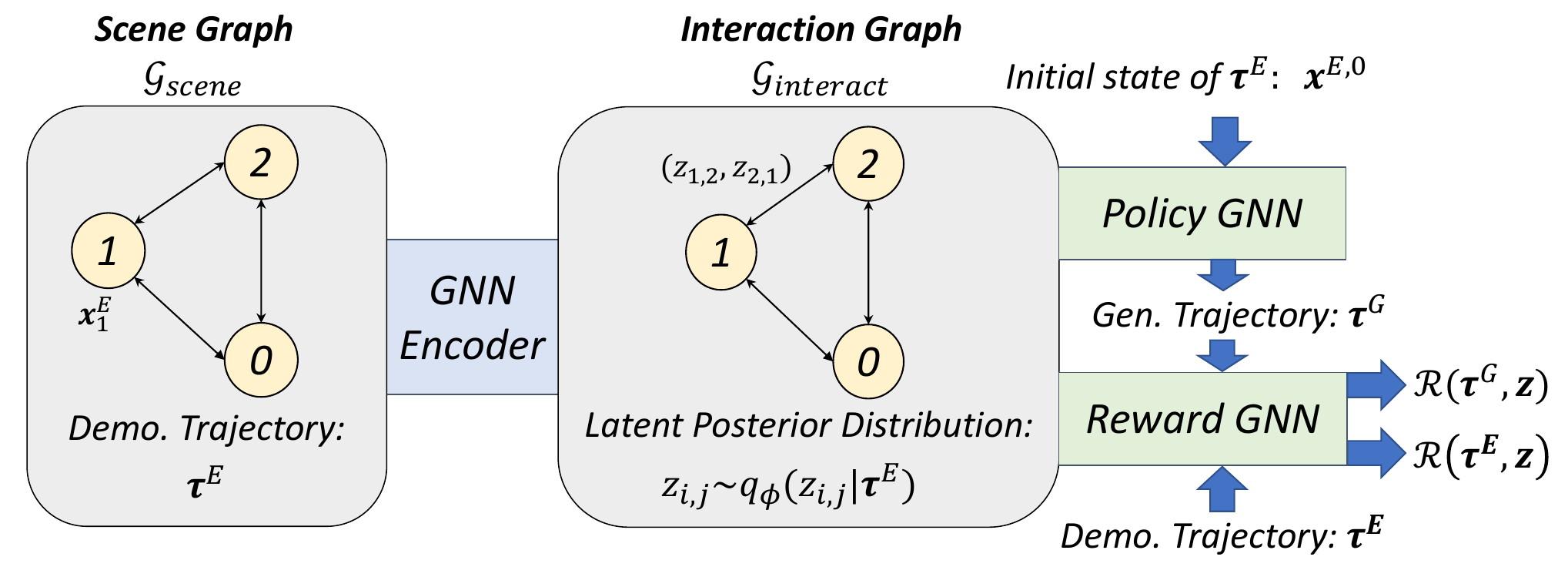} 
	\caption{Architecture of grounded relational inference model. Given a demonstration trajectory $\boldsymbol{\tau}^\mathrm{E}\in\mathcal{D}^\mathrm{E}$, the encoder operates over $\mathcal{G}_\mathrm{scene}$ and approximates the distribution $p(\mathbf{z}\vert\boldsymbol{\tau}^\mathrm{E})$ with $q_\phi(\mathbf{z}\vert\boldsymbol{\tau}^\mathrm{E})$. The policy decoder operates over a $\mathcal{G}_\mathrm{interact}$ sampled from the inferred $q_\phi(\mathbf{z}\vert\boldsymbol{\tau}^\mathrm{E})$ and models the policy $\boldsymbol{\pi}_\eta \left(\mathbf{a}^t\vert{\mathbf{x}^t, \mathbf{z}}\right)$. Given the initial state of $\boldsymbol{\tau}^\mathrm{E}$, we sample a trajectory $\boldsymbol{\tau}^\mathrm{G}$ by sequentially sampling $\mathbf{a}^t$ from $\boldsymbol{\pi}_\eta \left(\mathbf{a}^t\vert{\mathbf{x}^t, \mathbf{z}}\right)$ and propagating the state. Finally, We use the reward GNN to compute the cumulative rewards of $\boldsymbol{\tau}^\mathrm{G}$ and $\boldsymbol{\tau}^\mathrm{E}$ conditioned on the sampled $\mathcal{G}_\mathrm{interact}$.} \label{fig:architect}
	\end{figure*}

    The overall model structure is illustrated in Fig.~\ref{fig:architect}. Given a demonstration trajectory $\boldsymbol{\tau}^\mathrm{E}\in\mathcal{D}^\mathrm{E}$, the encoder operates over $\mathcal{G}_\mathrm{scene}$ and approximates the posterior distribution $p(\mathbf{z}\vert\boldsymbol{\tau}^\mathrm{E})$ with $q_\phi(\mathbf{z}\vert\boldsymbol{\tau}^\mathrm{E})$. \textcolor{black}{Following NRI, we parameterize the distribution of each edge latent variable with a softmax function, e.g., $q_\phi(\mathbf{z}_{i,j}\vert\boldsymbol{\tau}^\mathrm{E})=\mathrm{softmax}(f_{\mathrm{enc},\phi}(\boldsymbol{\tau}^\mathrm{E})_{i,j})$, where $f_{\mathrm{enc},\phi}(\cdot)$ denotes the encoder GNN.} The policy decoder operates over a $\mathcal{G}_\mathrm{interact}$ sampled from the inferred $q_\phi(\mathbf{z}\vert\boldsymbol{\tau}^\mathrm{E})$ and models the policy $\boldsymbol{\pi}_\eta \left(\mathbf{a}^t\vert{\mathbf{x}^t, \mathbf{z}}\right)$. \textcolor{black}{The policy distribution is parameterized as a diagonal Gaussian distribution whose mean and variance are generated by the decoder GNN.} Given an initial state, we can generate a trajectory by sequentially sampling $\mathbf{a}^t$ from $\boldsymbol{\pi}_\eta \left(\mathbf{a}^t\vert{\mathbf{x}^t, \mathbf{z}}\right)$ and propagating the state. The state is propagated with either the transition operator $\mathcal{T}$ if given, or a simulating environment if $\mathcal{T}$ is not accessible. We denote a generated trajectory given the initial state of $\tau^\mathrm{E}$ as $\tau^{\mathrm{G}}$. Since these two modules are essentially the same in NRI, we omit the detailed model structures here and include them in Appx.~\ref{app:model}.

    The reward decoder computes the reward of a state-action pair given the sampled edge variables. We use it to compute the cumulative rewards of $\boldsymbol{\tau}^\mathrm{G}$ and $\boldsymbol{\tau}^\mathrm{E}$ conditioned on the sampled $\mathcal{G}_\mathrm{interact}$. The reward decoder is in the form of Eqn. (\ref{eqn:reward}). Additionally, we augment the functions $r^n_\xi$ and $r^{e,k}_{\psi_k}$ with MLP shaping terms to mitigate the reward shaping effect~\cite{fu2017learning}, resulting in:
    \begin{equation}
        f^n_{\xi,\omega}(\mathbf{x}^t_j, \mathbf{a}^t_j, \mathbf{x}^{t+1}_j) = r^n_{\xi}(\mathbf{x}^t_j, \mathbf{a}^t_j)+h^n_\omega(\mathbf{x}^{t+1}_j)-h^n_\omega(\mathbf{x}^t_j), \label{eqn:node_reward}
    \end{equation}
    and
    \begin{equation}
    \begin{split}
        & f^{e,k}_{\psi_k, \chi_k}(\mathbf{x}^t_i, \mathbf{a}^t_i, \mathbf{x}^{t+1}_i, \mathbf{x}^t_j, \mathbf{a}^t_j, \mathbf{x}^{t+1}_j) = r^{{e},k}_{\psi_k}(\mathbf{x}^t_i, \mathbf{a}^t_i, \mathbf{x}^t_j, \mathbf{a}^t_j)\\
        &\quad \quad \quad + h^{e,k}_{\chi_k}(\mathbf{x}^{t+1}_i, \mathbf{x}^{t+1}_j)-h^{e,k}_{\chi_k}(\mathbf{x}^{t}_i, \mathbf{x}^{t}_j), \label{eqn:edge_reward}
    \end{split}
    \end{equation}
    where $h^n_{\omega}$ and $h^{e,k}_{\chi_k}$ are MLPs with parameters denoted by $\omega$ and $\chi$ respectively. We denote the shaped reward function of agent $v_j$ by $\mathbf{f}_{\xi,\omega,\psi,\chi}\left(\mathbf{x}^t, \mathbf{a}^t,\mathbf{x}^{t+1},\mathbf{z}\right)$, which equals to the left hand side of Eqn. (\ref{eqn:reward}) but with $r^n_\xi$ and $r^{e,k}_{\psi_k}$ substituted by the augmented rewards. The shaped reward function, together with the policy model, defines the discriminator which distinguishes $\boldsymbol{\tau}^\mathrm{G}$ from $\boldsymbol{\tau}^\mathrm{E}$:
    \begin{equation*}
    \begin{split}
        &\mathcal{D}_{\xi, \omega, \psi, \chi, \eta}(\mathbf{x}^t, \mathbf{a}^t, \mathbf{x}^{t+1}, \mathbf{z}) \\ 
        &\quad\quad\quad =\frac{\exp\left\{\mathbf{f}_{\xi,\omega,\psi,\chi}\left(\mathbf{x}^t, \mathbf{a}^t,\mathbf{x}^{t+1},\mathbf{z}\right)\right\}}{\exp\left\{\mathbf{f}_{\xi,\omega,\psi,\chi}\left(\mathbf{x}^t, \mathbf{a}^t, \mathbf{x}^{t+1}, \mathbf{z}\right)\right\}+\boldsymbol{\pi}_\eta\left(\mathbf{a}^t\vert \mathbf{x}^t, \mathbf{z}\right)}.
    \end{split}
    \end{equation*}
    
    \subsection{Training}
    We aim to train the three modules simultaneously. Consequently, we incorporate the encoder model $q_\phi\left(\mathbf{z}\vert \boldsymbol{\tau}^\mathrm{E}\right)$ into the objective function of AIRL, resulting in the optimization problem (\ref{eqn:opt-2}). The encoder is integrated into the minimization problem because the reward function directly depends on the latent space. The model is then trained by solving the problem (\ref{eqn:opt-2}) in an adversarial scheme: we alternate between training the encoder and reward for the minimization problem and training the policy for the maximization problem. Specifically, the objective for the encoder and reward is the following minimization problem given fixed $\eta$: 
    \begin{equation}
    \begin{aligned}
        \min_{\xi, \omega, \psi, \chi, \phi}\quad & \mathcal{J}(\xi, \omega, \psi, \chi, \phi, \eta) \\
        \textrm{s.t.}\quad & \mathbb{E}\left\{D_{KL}\left[q_\phi\left(\mathbf{z}\vert \boldsymbol{\tau}^\mathrm{E}\right))\vert\vert p(\mathbf{z})\right]\right\}\leqslant I_c. \label{eqn:min}
    \end{aligned}
    \end{equation}
    The objective for the policy is maximizing $\mathcal{J}(\xi, \omega, \psi, \chi, \phi, \eta)$ with fixed $\xi, \omega, \psi, \chi$ and $\phi$.

    \begin{figure*}[t]
    \begin{equation}
    \begin{split}
        \max_\eta \min_{\xi, \omega, \psi, \chi, \phi}\quad & \mathcal{J}(\xi, \omega, \psi, \chi, \phi, \eta)=\mathbb{E}_{\boldsymbol{\tau}^\mathrm{E}\sim\boldsymbol{\pi}^\mathrm{E}(\boldsymbol{\tau})}\Bigg\{\mathbb{E}_{\mathbf{z}\sim{q_\phi\left(\mathbf{z}\vert \boldsymbol{\tau}^\mathrm{E}\right)}}\bigg[-\sum_{t=0}^{T-1}\log \mathcal{D}_{\xi,\omega,\psi,\chi,\eta}(\mathbf{x}^{\mathrm{E}, t}, \mathbf{a}^{\mathrm{E},t}, \mathbf{x}^{\mathrm{E}, t+1},\mathbf{z}) \\
        &\qquad\qquad\qquad\quad\  -\mathbb{E}_{\boldsymbol{\tau}^\mathrm{G}\sim\boldsymbol{\pi}_\eta(\boldsymbol{\tau}\vert \mathbf{z})}\sum_{t=0}^{T-1}\log \left(1-\mathcal{D}_{\xi, \omega, \psi, \chi, \eta}(\mathbf{x}^{\mathrm{G}, t}, \mathbf{a}^{\mathrm{G},t}, \mathbf{x}^{\mathrm{G}, t+1}, \mathbf{z})\right)\bigg]\Bigg\}, \\
        \textrm{s.t.}\quad & \mathbb{E}_{\boldsymbol{\tau}^\mathrm{E}\sim\boldsymbol{\pi}^\mathrm{E}(\boldsymbol{\tau})}\left\{D_{KL}\left[q_\phi\left(\mathbf{z}\vert \boldsymbol{\tau}^\mathrm{E}\right))\vert\vert p(\mathbf{z})\right]\right\}\leqslant I_c, \label{eqn:opt-2}
    \end{split}
    \end{equation}
    \hrulefill
    \end{figure*}

    The objective function in the problem (\ref{eqn:opt-2}) is essentially the expectation of the objective function in the problem (\ref{eqn:opt}) over the inferred posterior distribution $q_\phi\left(\boldsymbol{z}\vert \boldsymbol{\tau}^\mathrm{E}\right)$ and the demonstration distribution $\boldsymbol{\pi}^\mathrm{E}\left(\boldsymbol{\tau}\right)$. The constraint enforces an upper bound $I_c$ on the KL-divergence between $q_\phi\left(\mathbf{z}\vert\boldsymbol{\tau}^\mathrm{E}\right)$ and the prior distribution $p(\mathbf{z})$. A sparse prior is chosen to encourage sparsity in $\mathcal{G}_\mathrm{interact}$. It has a similar regularization effect as the $D_{KL}$ term in ELBO. We borrow its format from variational discriminator bottleneck (VDB)~\cite{peng2018variational}. VDB improves adversarial training by constraining the information flow from the input to the discriminator. The KL-divergence constraint is derived as a variational approximation to the information bottleneck~\cite{alemi2016deep}. Although having different motivations, we adopt it for two reasons. First, the proposed model is not generative because our goal is not synthesizing trajectories from the prior $p(\mathbf{z})$, but inferring the posterior $p\left(\mathbf{z}\vert\boldsymbol{\tau}^\mathrm{E}\right)$. Therefore, regularization derived from information bottleneck is more sensible compared to ELBO. Second, the constrained problem (\ref{eqn:min}) can be relaxed by introducing a Lagrange multiplier $\beta$. During training, $\beta$ is updated through dual gradient descent as follows:
    \begin{equation}
        \beta \leftarrow \max\left(0, \alpha_\beta\left(\mathbb{E}\left\{D_{KL}\left[q_\phi\left(\mathbf{z}\vert \boldsymbol{\tau}^\mathrm{E}\right))\vert\vert p(\mathbf{z})\right]\right\}- I_c \right) \right) \label{eqn:adapt}
    \end{equation}
    We find the adaptation scheme particularly advantageous. The model can focus on inferring $\mathbf{z}$ for reward learning after satisfying the sparsity constraint because the magnitude of $\beta$ decreases towards zero once the constraint is satisfied. However, it is worth noting that our framework does not rely on the bottleneck constraint to induce a semantically meaningful latent space as in~\cite{higgins2016beta}. In contrast, GRI relies on structured reward functions to ground the latent space into semantic interactive behaviors. The bottleneck serves as a regularization to find out the minimal interaction graph to represent the interactions. In fact, we trained the baseline NRI models with the same constraints and weight update scheme. The experimental results show that the constraint itself is not sufficient to induce a sparse interaction graph.
    
   In general, when the dynamics $\mathcal{T}$ is unknown or non-differentiable, maximum entropy RL algorithms~\cite{levine2018reinforcement} are adopted to optimize the policy. In this work, we assume known and differentiable dynamics, which is a reasonable assumption for the investigated scenarios. It allows us to directly backpropagate through the trajectory for gradient estimation, which simplifies the training procedure.
    

\section{Experiments}
\label{sec:experiments}
We evaluate the proposed GRI model on a synthetic dataset as well as a naturalistic traffic dataset. The synthetic data are generated using policy models trained given the ground-truth reward function and interaction graph. We intend to verify if GRI can induce a semantically meaningful relational latent space and infer the underlying relations precisely. The naturalistic traffic data are extracted from the NGSIM dataset. We aim to validate if GRI can model real-world traffic scenarios effectively with the grounded latent space. Unlike synthetic agents, we do not have the privilege to access the ground-truth graphs governing human drivers' interactions. Instead, we construct hypothetical graphs after analyzing the segmented data. The hypotheses reflect humans' understanding of traffic scenarios. \textcolor{black}{Moreover, the hypothetical graphs are built upon a set of interactive behavior whose characteristics are described by the designed reward functions. We would like to see if the reward functions can incorporate the semantic information into the latent space, and let GRI model real-world interactive systems in the same way as humans.} In each setting, we consider two scenarios, car-following, and lane-changing.

\subsection{Baselines} \label{sec:baseline}
The main question of interest is whether GRI can induce semantically meaningful interaction graphs. To answer the question, the most important baseline model for comparison is NRI, because GRI shares the same prior distribution of latent variables with NRI. Comparing the posterior distributions provides insights into whether the structured reward functions can ground the latent space into semantic interactive behaviors. In each experiment, the baseline NRI model has the same encoder and policy decoder as the GRI model. Besides, as stated in Sec.~\ref{sec:method}, the same bottleneck constraint and the weight update scheme in Eqn. (\ref{eqn:adapt}) were applied as regularization for minimal representation. 

Another model for comparison is a supervised policy decoder. We assume that ground-truth graphs or human hypotheses are available. Therefore, we can directly train a policy decoder in a supervised way. The ground-truth graph is fed to the policy decoder as a substitute for the interaction graph sampled from the encoder output $q_\phi(\mathbf{z}\vert\boldsymbol{\tau}^\mathrm{E})$. The training of the decoder becomes a simple regression problem. We used mean square error as the loss function to train it. 

As additional information is granted, it is unfair to directly compare the performance of GRI with the supervised policy model. Since the supervised model is trained with the ground-truth interaction graphs governing the systems, it is expected to achieve smaller reconstruction errors. However, the supervised baseline provides some useful insights. In the naturalistic traffic scenarios, the supervised model gives us some insights into whether the human hypotheses are reasonable. If the supervised model can reconstruct the trajectories precisely, it will justify our practice to adopt graph accuracy as one of the evaluation metrics. 

\textcolor{black}{More importantly, in Sec.~\ref{sec:ood}, we demonstrate that GRI's latent space still maintains its semantic meaning under some perturbations to the initial states, whereas the decoders of baseline models fail to synthesize those behaviors under the same perturbations, including the supervised policy decoder which is trained with the ground-truth interaction graphs. It supports our argument that direct supervision via interaction labels is not sufficient to guide the policy to synthesize behaviors with correct semantic meaning.}

There exist other alternatives for the purpose of trajectory reconstruction. However, it is not our goal in this paper to find an expressive model for accurate reconstruction. Therefore, we do not consider other baselines from this perspective. For the task of grounding the latent space into semantic interactive driving behaviors, we did not find any exact alternatives in the literature. \textcolor{black}{For the specific scenarios studied in this paper, we may design some rule-based approaches to directly infer the interaction graph. However, it is difficult to decide the parameters that best describe the interactive behaviors, because there is a spectrum in how people follow the rules~\cite{8814167}. In this paper, we are interested in a data-driven module that can be incorporated into an end-to-end learning model and has the potential to be generalized to complicated driving scenarios and systems in other domains. Apart from GRI, a potential alternative solution could be adopting a differentiable logic module. For instance, Leung et al.~\cite{8814167} proposed a differentiable parametric Signal Temporal Logic formula (pSTL) which could be learned from data. We will investigate this direction in our future works.}

\begin{figure*}[t]
\centering
\includegraphics[width=6.5in]{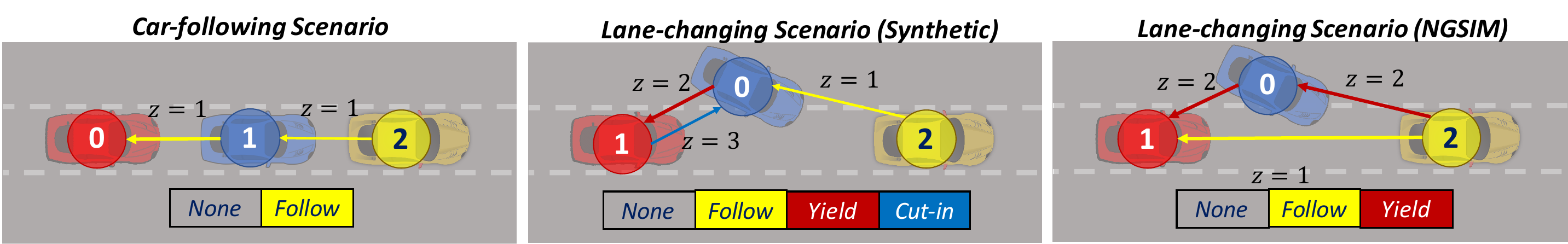} 
\caption{Test scenarios with the underlying interaction graphs. In the synthetic scenarios, the graphs are the ground-truth ones governing the synthetic experts. In the naturalistic traffic scenarios, the graphs are human hypotheses reflecting humans' understanding of the traffic scenarios.} \label{fig:scene}
\end{figure*}

\subsection{Evaluation Metrics}
To evaluate a trained model, we sample a $\boldsymbol{\tau}^\mathrm{E}$ from the test dataset and extract the maximum posterior probability (MAP) estimate of edge variables, $\hat{\mathbf{z}}$, from $q_\phi(\mathbf{z}\vert\boldsymbol{\tau}^\mathrm{E})$. Afterward, we obtain a single sample of trajectories $\hat{\boldsymbol{\tau}}$ by executing the mean value of the policy output. The root mean square errors (RMSE) of states and the accuracy of $\mathcal{G}_\mathrm{interact}$ are selected as the evaluation metrics, which are computed based on $\hat{\mathbf{z}}$, $\hat{\boldsymbol{\tau}}$, $\boldsymbol{\tau}^\mathrm{E}$, and the ground truth or hypothetical latent variables denoted by $\mathbf{z}^\mathrm{E}$:
\begin{equation*}
\begin{split}
    \mathrm{RMSE}_\epsilon &= \sqrt{\frac{1}{(N-1)T}\sum_{j=1}^{N}\sum_{t=0}^{T-1}(\epsilon^{\mathrm{E},t}_j-\hat{\epsilon}^t_j)^2}, \\
    \mathrm{Accuracy} &= \frac{\sum_{i=1}^{N} \sum_{j=1, j\neq i}^{N}\mathbf{1}(z^{\mathrm{E}}_{i,j}=\hat{z}_{i,j})}{N(N-1)}.
\end{split}
\end{equation*}
If multiple edge types exist, we test all the possible permutations of edge types and report the one with the highest graph accuracy for NRI.

\textcolor{black}{It is worth noting that the graph accuracy on the naturalistic traffic dataset merely quantifies the divergence between the inferred graphs and the hypotheses we construct. We anticipate that GRI can attain a higher accuracy than NRI. It will imply that we can incorporate human domain knowledge into GRI and induce a semantic relational latent space consistent with the hypotheses built upon the same domain knowledge. However, a low graph accuracy does not necessarily mean that humans cannot interpret the inferred graphs well. The hypothetical graphs represent one perspective to interpret the interactive scenes. It is possible that NRI may find another sensible way to categorize and interpret the interactions, which can also be understood by humans.}

\textcolor{black}{To further study the explainability of the learned latent spaces, we want to look into the inferred graphs and have a qualitative comparison between the latent spaces learned by the two models. For each setting, we compute the distribution of estimated edge variables $\hat{\mathbf{z}}$ over the test dataset. As in~\cite{kipf2018neural}, we visualize the results in multiple adjacency matrices corresponding to different edge types. In the adjacency matrix corresponding to the $k^\mathrm{th}$ type of interaction, the element $A_{i,j}$ indicates the relative frequency of $\hat{z}_{j,i}=k$, where $\hat{z}_{j,i}$ is the latent variable for the edge from node $j$ to node $i$. In other words, $A_{i,j}$ equals the ratio of test samples where the model infers $\hat{z}_{j,i}=k$. By inspecting the edge-type distributions, we can get some extra insights into the explainability of the two models beyond the quantitative metrics.}

\subsection{Synthetic Scenes}\label{sec:synthetic}

\begin{table*}[t]
\centering
\caption{Performance Comparison on Synthetic Dataset}
\label{table:synthetic}
\begin{threeparttable}
\centering
\begin{tabular}{|c|c|c|c|c|c|c|c|}
\hline
\multirow{2}{*}{Model} & \multicolumn{3}{c|}{Car Following ($\Delta t=0.2s$, $T=20$)} & \multicolumn{4}{c|}{Lane Changing ($\Delta t=0.2s$, $T=30$)} \\ \cline{2-8} 
& $\mathrm{RMSE}_x(\mathrm{m})$ & $\mathrm{RMSE}_v(\mathrm{m/s})$ &  $\mathrm{Accuracy}(\%)$ & $\mathrm{RMSE}_x(\mathrm{m})$ & $\mathrm{RMSE}_y(\mathrm{m})$ & $\mathrm{RMSE}_v(\mathrm{m/s})$ & $\mathrm{Accuracy}(\%)$   \\ \hline
GRI & $0.241\pm{0.125}$ & $0.174\pm{0.068}$ & $\mathbf{100.00\pm{0.00}}$ & \textcolor{black}{$0.529\pm{0.230}$} & \textcolor{black}{$0.207\pm{0.046}$} & \textcolor{black}{$0.303\pm{0.128}$} & \textcolor{black}{$\mathbf{99.95\pm{0.01}}$}       \\ \hline
NRI & ${0.047\pm{0.024}}$ & ${0.056\pm{0.015}}$ & $66.70\pm{0.00}$ & \textcolor{black}{${0.109\pm{0.045}}$} & \textcolor{black}{${0.155\pm{0.038}}$} & \textcolor{black}{${0.061\pm{0.016}}$} & \textcolor{black}{$55.9\pm{7.98}$}     \\ \hline
Supervised & $\mathbf{0.039\pm{0.016}}$ & $\mathbf{0.050\pm{0.009}}$ & - & \textcolor{black}{$\mathbf{0.062\pm{0.027}}$} & \textcolor{black}{$\mathbf{0.145\pm{0.035}}$} & \textcolor{black}{$\mathbf{0.048\pm{0.011}}$} & -\\ \hline
\end{tabular}
\begin{tablenotes}
	\item[1] The data is presented in form of $\text{mean}\pm{\text{std}}$.
\end{tablenotes}
\end{threeparttable}
\end{table*}

As mentioned above, we designed two synthetic scenarios, car-following and lane-changing. The two scenes and their underlying interaction graphs are illustrated in Fig.~\ref{fig:scene}. In both scenarios, we have a leading vehicle whose behavior does not depend on the others. Its trajectory is given without the need for reconstruction. We assume it runs at constant velocity. The other vehicles interact with each other and the leader in different ways. In the car-following scene, we model the system with two types of edges: $z_{i,j}=1$ means that Vehicle $j$ follows Vehicle $i$; $z_{i,j}=0$ means that Vehicle $j$ does not interact with Vehicle $i$. In the lane-changing scene, two additional edge types are introduced: $z_{i,j}=2$ means that Vehicle $j$ yields to Vehicle $i$; $z_{i,j}=3$ means that Vehicle $j$ cuts in front of Vehicle $i$.

The MDPs for the tested scenarios are specified as follows. In the car-following scene, since the vehicles mainly interact in the longitudinal direction, we only model their longitudinal dynamics to simplify the problem. For all $j\in\{1,2,3\}$, the state vector of Vehicle $j$ consists of three states: $\mathbf{x}^t_{j}=\left[x^t_j\ v^t_j\ a^t_j\right]^\intercal$, where $x^{t}_j$ is the longitudinal coordinate, $v^{t}_j$ is the velocity, and $a^t_j$ is the acceleration. There is only one control input which is the jerk. We denote it as $\delta a^t_j$. The dynamics are governed by a 1D point-mass model: 
\begin{align*}
x^{t+1}_j &= x^t_j + v^t_j\Delta t + \frac{1}{2}a^t_j{\Delta t}^2,\\
v^{t+1}_j &= v^t_j + a^t_j\Delta t, \\
a^{t+1}_j &= a^t_j + \delta a^t_j\Delta t,
\end{align*}
where $\Delta t$ is the sampling time. \textcolor{black}{Note that the dynamics of $x^{t}_j$ is discretized with zero-order hold discretization so that the longitudinal acceleration can directly affect $x^{t}_j$ without delay. In practice, we find GRI benefits from reducing the control delay as the longitudinal coordinates under different policies will diverge at earlier time steps. It is then easier for GRI to infer the latent interaction types.} In the lane-changing scene, we consider both longitudinal and lateral motions. The state vector consists of six states instead: $\mathbf{x}^t_{j}=\left[x^t_j\ y^t_j\ v^t_j\ \theta^t_j\ a^t_j\ \omega^t_j\right]^\intercal$. The three additional states are the lateral coordinate $y^{t}_j$, the yaw angle $\theta^t_j$, and the yaw rate $\omega^t_j$. There is one additional action which is the yaw acceleration, denoted by $\delta \omega^t_j$. We model the vehicle as a Dubins' car:
\begin{align*}
x^{t+1}_j &= x^t_j + v^t_j\cos(\theta^t_j)\Delta t, \\
y^{t+1}_j &= y^t_j + v^t_j\sin(\theta^t_j)\Delta t, \\
v^{t+1}_j &= v^t_j + a^t_j\Delta t, \\
\theta^{t+1}_j &= \theta^{t}_j + \omega^t_j\Delta t, \\
a^{t+1}_j &= a^t_j + \delta a^t_j\Delta t, \\
\omega^{t+1}_j &= \omega^{t}_j + \delta \omega^t_j\Delta t.
\end{align*}

The structured reward functions were designed based on expert domain knowledge (e.g. transportation studies~\cite{kesting2010enhanced, treiber2000congested}). We mainly referred to~\cite{sun2018probabilistic, naumann2020analyzing} in this paper. For the car-following behavior, its reward function is defined as follows:
\begin{equation*}
\begin{split}
    r^{e,1}_{\psi_1} \left(\mathbf{x}^t_i, \mathbf{x}^t_j\right) = & - \left(1+\exp(\psi_{1,0})\right) g_\mathrm{IDM}(\mathbf{x}^t_i, \mathbf{x}^t_j) \\
	& - \left(1+\exp(\psi_{1,1})\right) g_\mathrm{dist}(\mathbf{x}^t_i, \mathbf{x}^t_j) \\
	& - \left(1+\exp(\psi_{1,2})\right) g_\mathrm{lat}(\mathbf{x}^t_i, \mathbf{x}^t_j),
\end{split}
\end{equation*}
where the features are defined as:
\begin{align}
g_\mathrm{IDM}(\mathbf{x}^t_i, \mathbf{x}^t_j) & = \left(\max{\left(x^t_i-x^t_j, 0\right)}-\Delta x^{\mathrm{IDM},t}_{i,j}\right)^2, \label{eqn:fIDM}\\
g_\mathrm{dist}(\mathbf{x}^t_i, \mathbf{x}^t_j) & = \exp\left(-\frac{\left(\max{\left(x^t_i-x^t_j, 0\right)}\right)^2}{\zeta^2}\right), \label{eqn:fdist}\\
g_\mathrm{lat}(\mathbf{x}^t_i, \mathbf{x}^t_j) & =\left(y^t_j - g_\mathrm{center}(y^t_i)\right)^2.\nonumber
\end{align}
The feature $g_\mathrm{IDM}$ suggests a spatial headway $\Delta x^{\mathrm{IDM},t}_{i,j}$ derived from the intelligent driver model (IDM)~\cite{kesting2010enhanced}. The feature $f_\mathrm{dist}$ ensures a minimum collision-free distance. We penalize the following vehicle for surpassing the preceding one with the help of $x^{\mathrm{IDM},t}_{i,j}$ in Eqn. (\ref{eqn:fIDM}) and Eqn. (\ref{eqn:fdist}). The last feature $g_\mathrm{lat}$ exists only in lane-changing. It regulates the following vehicle to stay in the same lane as the preceding one with the help of $g_\mathrm{center}$, which determines the lateral coordinate of the corresponding centerline based on the position of the preceding vehicle. \textcolor{black}{Altogether, the features define the following behavior as staying in the same lane as the preceding vehicle whereas keeping a safe longitudinal headway.}

The reward function for yielding is defined as:
\begin{equation*}
\begin{split}
r^{e,2}_{\psi_2} \left(\mathbf{x}^t_i, \mathbf{x}^t_j\right) = & - \left(1+\exp(\psi_{2,0})\right) g_\mathrm{yield}(\mathbf{x}^t_i, \mathbf{x}^t_j) \\
& - \left(1+\exp(\psi_{2,1})\right) g_\mathrm{dist}(\mathbf{x}^t_i, \mathbf{x}^t_j).
\end{split}
\end{equation*}
The feature $g_\mathrm{dist}$ is defined in Eqn. (\ref{eqn:fdist}). The other feature $g_\mathrm{yield}$ suggests an appropriate spatial headway for yielding:
\begin{align}
g_\mathrm{yield}(\mathbf{x}^t_i, \mathbf{x}^t_j) = & \mathbf{1}\left(g_\mathrm{center}(y^t_j)=g_\mathrm{center}(y^t_i)\right)g_\mathrm{IDM}(\mathbf{x}^t_i, \mathbf{x}^t_j)\nonumber \\
+ & \mathbf{1}\left(g_\mathrm{center}(y^t_j)\neq g_\mathrm{center}(y^t_i)\right)g_\mathrm{goal}(\mathbf{x}^t_i, \mathbf{x}^t_j), \nonumber\\
g_\mathrm{goal}(\mathbf{x}^t_i, \mathbf{x}^t_j) = & \left(\max{\left(x^t_i-x^t_j-\Delta x^{\mathrm{yield}}, 0\right)}\right)^2. \label{eqn:fgoal} 
\end{align}
The suggested headway is set to be a constant value, $\Delta x^{\mathrm{yield}}$, when the other vehicle is merging, and switches to $\Delta x^{\mathrm{IDM},t}_{i,j}$ once the merging vehicle enters into the same lane, where its behavior becomes consistent with car-following. \textcolor{black}{We follow~\cite{sun2018probabilistic} to adopt different reward functions depending on the lanes where the vehicles are located. Merging occurs during a short period of time. Therefore, we assume the driver sets a fixed short-term goal distance as in~\cite{sun2018probabilistic} and then transits to the following behavior afterward.}

The reward function for cutting-in is quite similar:
\begin{equation*}
\begin{split}
r^{e,3}_{\psi_3} \left(\mathbf{x}^t_i, \mathbf{x}^t_j\right) = & - \left(1+\exp(\psi_{3,0})\right) g_\mathrm{goal}(\mathbf{x}^t_j, \mathbf{x}^t_i) \\
& - \left(1+\exp(\psi_{3,1})\right) g_\mathrm{dist}(\mathbf{x}^t_j, \mathbf{x}^t_i),
\end{split}
\end{equation*}
where the features are defined as in Eqn. (\ref{eqn:fdist}) and Eqn. (\ref{eqn:fgoal}), but with the input arguments switched, because the merging vehicle should stay in front of the yielding one. 

Apart from the edge rewards, all the agents share the same node reward function. The following one is adopted for lane-changing:
\begin{equation*}
	\begin{split}
	r^n_\xi (\mathbf{x}^t_j, \mathbf{a}^t_j)=&-\left(1+\exp(\xi_0)\right)f_v(\mathbf{x}_j^t)\\
	&-\left(1+\exp({\xi}_{1:3})\right)^\intercal {f}_\mathrm{state}(\mathbf{x}^t_j)\\
	&-\left(1+\exp({\xi}_{4:5})\right)^\intercal {f}_\mathrm{action}(\mathbf{a}^t_j)\\
	&-\left(1+\exp(\xi_{6})\right)f_\mathrm{lane}(\mathbf{x}_j^t),
	\end{split}
\end{equation*}
where $f_\mathrm{state}$ and $f_\mathrm{action}$ take the element-wise square of $\left[a^t_j\ \theta^t_j\  \omega^t_j \right]$ and $\left[\delta a^t_j\ \delta\omega^t_j \right]$ respectively. \textcolor{black}{It penalizes large control inputs as well as drastic longitudinal and angular motions to induce smooth and comfortable maneuvers.} The feature $f_v$ is the squared error between $v^t_j$ and the speed limit $v_\mathrm{lim}$. \textcolor{black}{It regulates the vehicles to obey the speed limit.} The last term $f_\mathrm{lane}$ penalizes the vehicle for staying close to the lane boundaries. For car-following, we simply remove those terms that are irrelevant in 1D motion. 
In all the reward functions, the parameters collected in $\psi$ and $\xi$ are unknown during training and inferred by GRI. We take the exponents of them and add one to the results. It enforces the model to use the features when modeling the corresponding interactions. 

With the scenarios defined above, we aim to generate one dataset for each scenario. For each scenario, we randomly sampled the initial states of the vehicles and trained an expert policy given the ground-truth reward functions and the interaction graph. Afterward, we used the trained policy to generate the dataset. The same sampling scheme was used to sample the initial states.

{\bf Results.} On each dataset, we trained a GRI model with the policy decoder (\ref{eqn:policy1})-(\ref{eqn:policy3}) introduced in Appx.~\ref{app:model}. The results are summarized in Table~\ref{table:synthetic}. The NRI model can reconstruct the trajectories with errors close to the supervised policy. However, it learns a relational latent space that is different from the one governing the demonstration; Therefore, the edge variables cannot be interpreted as those semantic interactive behaviors. In contrast, our GRI model interprets the interactions consistently with the domain knowledge inherited in the demonstration and recovers the interaction graph with high accuracy. It has larger reconstruction errors compared to the baseline approaches. However, it still sufficiently recovers the interactive behaviors, and the reconstructed trajectories are sensible (see Appx.~\ref{app:visual}). 

\begin{figure}[t]
	\centering
	\includegraphics[width=3in]{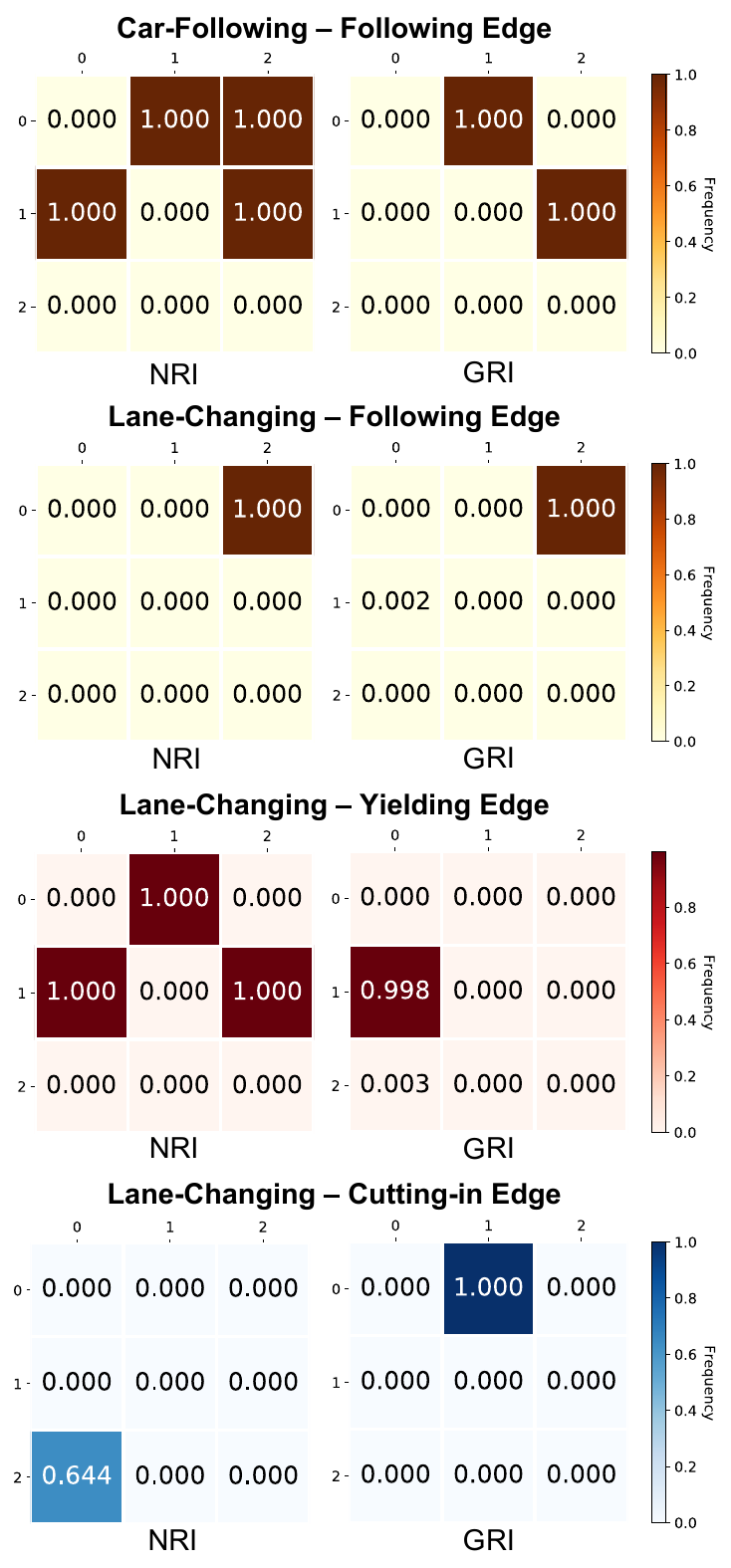}
	\caption{The empirical distribution of estimated edge variables $\hat{z}$ over the test dataset in the synthetic scenarios. We summarize the results in multiple adjacency matrices corresponding to different edge types. In the adjacency matrix corresponding to the $k^\mathrm{th}$ type of interaction, the element $A_{i,j}$ indicates the relative frequency of $\hat{z}_{j,i}=k$, where $\hat{z}_{j,i}$ is the latent variable for the edge from node $j$ to node $i$. } \label{fig:sim_graph}
    \vspace{-6pt}
\end{figure}

We computed the empirical distribution of the estimated edge variables $\hat{z}$ over the test dataset. The results are summarized in Fig.~\ref{fig:sim_graph}. The distribution concentrates on a single interaction graph for both models in both scenarios\textemdash as opposed to the case on the naturalistic traffic dataset introduced in the next section\textemdash because the synthetic agents have consistent interaction patterns over all the samples. \textcolor{black}{We observe that NRI learns symmetric relations: In both scenarios, the NRI model assigns the same edge types to the edges $e_{0,1}$ and $e_{1,0}$. It is difficult to interpret their semantic meaning because those pairwise interactions are asymmetric in our synthetic scenes. In contrast, the reward functions in our GRI model enforce an asymmetric relational latent space.}

\subsection{Naturalistic Traffic Scenes}
To evaluate the proposed method in real-world traffic scenarios, we investigated the same scenarios as in the synthetic case, car-following and lane-changing. we segmented data from the Highway-101 and I-80 datasets of NGSIM. Afterward, we further screened the data to select those interactive samples and ensure that no erratic swerving or multiple lane changes occurred. Unlike synthetic agents, human agents do not have a ground-truth interaction graph that governs their interactions. Instead, we constructed hypothetical $\mathcal{G}_\mathrm{interact}$ after analyzing the segmented data. The hypotheses for the two scenarios are depicted in Fig.~\ref{fig:scene}. The one for car-following is identical to the ground-truth interaction graph we designed for the synthetic agents. However, we proposed a different hypothesis for lane-changing. We excluded the cutting-in relation to reduce the number of edge types and therefore simplify the training procedure. Moreover, we differentiated distinct interactions according to the vehicle's lateral position. We say that a vehicle yields to its preceding vehicle if they drive in neighboring lanes, whereas it follows the preceding one if they drive in the same lane. 

\textcolor{black}{As in the synthetic scenes, the trajectory of the leading vehicle is given without the need for reconstruction. We feed the ground-truth state of the leading vehicle sequentially to the policy decoder when decoding the trajectories of the other vehicles. This practice enables us to heuristically isolate a small interacting group out of the numerous number of vehicles on the highway. While the leading vehicle's behavior depends on the other vehicles, it is fairly reasonable to assume that the behavior of the modeled following vehicles is independent of other surrounding vehicles on the road after conditioning on the trajectory of the leading vehicle. Even though there may still exist other surrounding vehicles interacting with them, their influence should be subtle. The models should be able to well capture the interactions among the modeled subset while marginalizing those subtle effects.}

The node dynamics are the same as in the synthetic scene for car-following. For lane-changing, since we did not have accurate heading information, we adopted a 2D point-mass model instead. Since the behavior of human drivers is much more complicated than the synthetic agents, we designed reward functions with larger model capacities using neural networks. In car-following, the reward functions are defined as follows:
\begin{align*}
    r^{e,1}_{\psi_1} \left(\mathbf{x}^t_i, \mathbf{x}^t_j\right) = & - \left(1+\exp(\psi_{1,0})\right) g_\mathrm{v}^\mathrm{NN}(\mathbf{x}^t_i, \mathbf{x}^t_j) \\
    & - \left(1+\exp(\psi_{1,1})\right) g_\mathrm{s}^\mathrm{NN}(\mathbf{x}^t_i,\mathbf{x}^t_j),\\
    r^n_\xi \left(\mathbf{x}^t_j, \mathbf{a}^t_j\right) = &-\left(1+\exp(\xi_0)\right)f_\mathrm{v}^\mathrm{NN}(\mathbf{x}^t_j) \\
    &- \left(1+\exp(\xi_1)\right)f_\mathrm{acc}(\mathbf{x}^t_j)\\
    &- \left(1+\exp(\xi_2)\right)f_\mathrm{jerk}(\mathbf{x}^t_j, \mathbf{a}^t_j),
\end{align*}
where the features are defined as:
\begin{align*}
    f_\mathrm{v}^\mathrm{NN}(\mathbf{x}^t_j) & = \left(v^t_{j}-h_1(\mathbf{x}^t_j)\right)^2,\\
    g_\mathrm{v}^\mathrm{NN}(\mathbf{x}^t_i, \mathbf{x}^t_j) & = \left(v^t_{j}-h_2(\mathbf{x}^t_i, \mathbf{x}^t_j)\right)^2,\\
    g_\mathrm{s}^\mathrm{NN}(\mathbf{x}^t_i,\mathbf{x}^t_j) & = \mathrm{ReLU} {\left(h_3\left(\mathbf{x}^t_i, \mathbf{x}^t_j\right)-x^t_i+x^t_j\right)}^2.
\end{align*}
The features $f_\mathrm{acc}$ and $f_\mathrm{jerk}$ penalize the squared magnitude of acceleration and jerk \textcolor{black}{to induce smooth and comfortable maneuver}. The functions $h_1$, $h_2$, and $h_3$ are neural networks with ReLU output activation. The feature $g^\mathrm{NN}_\mathrm{s}$ is the critical component that shapes the car-following behavior. It learns a non-negative reference headway and penalizes the following vehicle for violating it. The feature $g_\mathrm{v}^\mathrm{NN}$ and $f_\mathrm{v}^\mathrm{NN}$ suggest reference velocities considering interaction and merely itself respectively. \textcolor{black}{The edge reward function has a large modeling capacity because we let it learn adaptive reference headway and velocity from data. Nevertheless, it still defines the fundamental characteristic of the following behavior, which is always staying behind the preceding vehicle.}

\begin{figure}[t]
	\centering
	\includegraphics[height=1.1in]{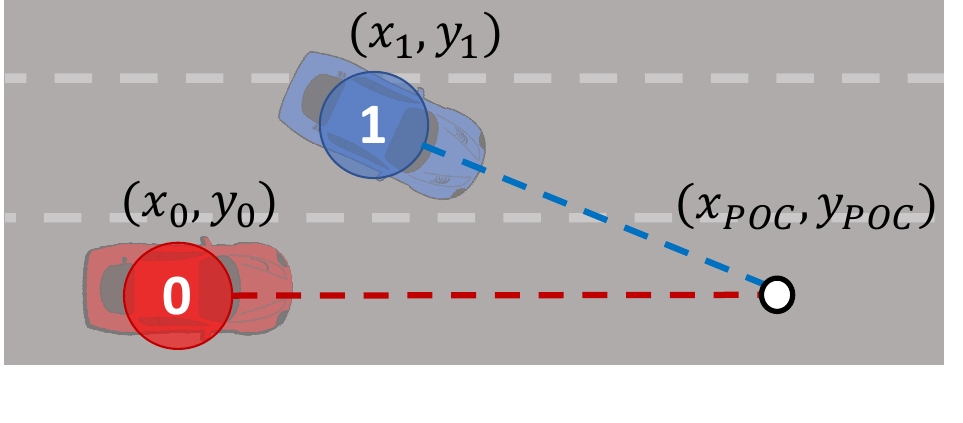} 
	\caption{Collision point diagram. At every timestep, the heading vector of the agents' can be calculated approximating the motion as linear. The intersection between these vectors is taken to be the collision point where the agents would collide if a yield action is not taken. } \label{fig:poc_viz}
\end{figure}

\textcolor{black}{In lane-changing, the node reward function and the edge reward function for the following behavior are similar to those in the car-following scenario. The node reward function has an additional term for the lateral position, which encourages the vehicles to drive on the target lane, i.e., the lane where the leading vehicle is driving. It also has additional terms to penalize the magnitude of lateral velocity and acceleration to induce comfortable maneuvers.} To design the yielding reward, we define a collision point of two vehicles based on their states. We approximate the vehicles' trajectories as piecewise-linear between sequential timesteps and compute the collision point as the intersection between their trajectories (Fig.~\ref{fig:poc_viz}). We threshold the point if it exceeds a hard-coded range of interest (e.g. if it is behind the vehicles or greater than a certain distance). Afterward, we define the distance-to-collision ($d_{poc}$) as the longitudinal distance from the vehicle to the collision point, and the time-to-collision ($T_{col}$) as the time to reach the collision point calculated by dividing $d_{poc}$ with the velocity of the vehicle. Then the yielding reward function is defined as follows:
\begin{equation*}
\begin{split}
r^{e,2}_{\psi_2} \left(\mathbf{x}^t_i, \mathbf{x}^t_j\right) = & - \left(1+\exp(\psi_{2,0})\right)g_\mathrm{spatial}^\mathrm{NN}(\mathbf{x}^t_i, \mathbf{x}^t_j) \\
& - \left(1+\exp(\psi_{2,1})\right)g_\mathrm{time}^\mathrm{NN}(\mathbf{x}^t_i, \mathbf{x}^t_j),
\end{split}
\end{equation*}
where
\begin{equation*}
\begin{split}
g_\mathrm{spatial}^\mathrm{NN}(\mathbf{x}^t_i, \mathbf{x}^t_j) &= \mathrm{ReLU}{\left((x_{j}-x_{poc})-h_\mathrm{d_{poc}}(\mathbf{x}^t_i, \mathbf{x}^t_j)\right)}^2,\\
g_\mathrm{time}^\mathrm{NN}(\mathbf{x}^t_i, \mathbf{x}^t_j) &= 
\mathrm{ReLU}{\left(h_\mathrm{T_{col}}(\mathbf{x}^t_i, \mathbf{x}^t_j)-(T_{col_{i}}-T_{col_{j}})\right)}^2.
\end{split}
\end{equation*}
The functions $h_\mathrm{d_{poc}}$ and $h_\mathrm{T_{col}}$ are neural networks with ReLU output activation. The $g_\mathrm{spatial}$ term learns a spatial aspect of the yield behavior and compares the agent's distance from the estimated collision point with the NN-learned \textit{safe} reference within which the lane-changing maneuver can be done. The second term $g_\mathrm{time}$ adds a temporal aspect, by enforcing the vehicle to ensure a minimum \emph{safe} time headway. \textcolor{black}{We adopt $g_\mathrm{time}$ because time-to-collision is an important measure in traffic safety assessment~\cite{minderhoud2001extended}}. The intuition behind this is to ensure that the vehicles do not occupy the same position at the same time.

{\bf Results.} For each scenario, we trained a GRI model with the recurrent policy decoder (\ref{eqn:rnn1})-(\ref{eqn:rnn4}) in Appx.~\ref{app:model}. The results are summarized in Table~\ref{table:ngsim}. In car-following, the NRI model still performs better on trajectory reconstruction, but the GRI model achieves comparable RMSE on NGSIM dataset. In lane-changing, their comparison is consistent: The NRI model slightly outperforms our model in trajectory reconstruction; Our model dominates the NRI model in graph accuracy.

\begin{figure}[t]
	\centering
	\includegraphics[width=2.9in]{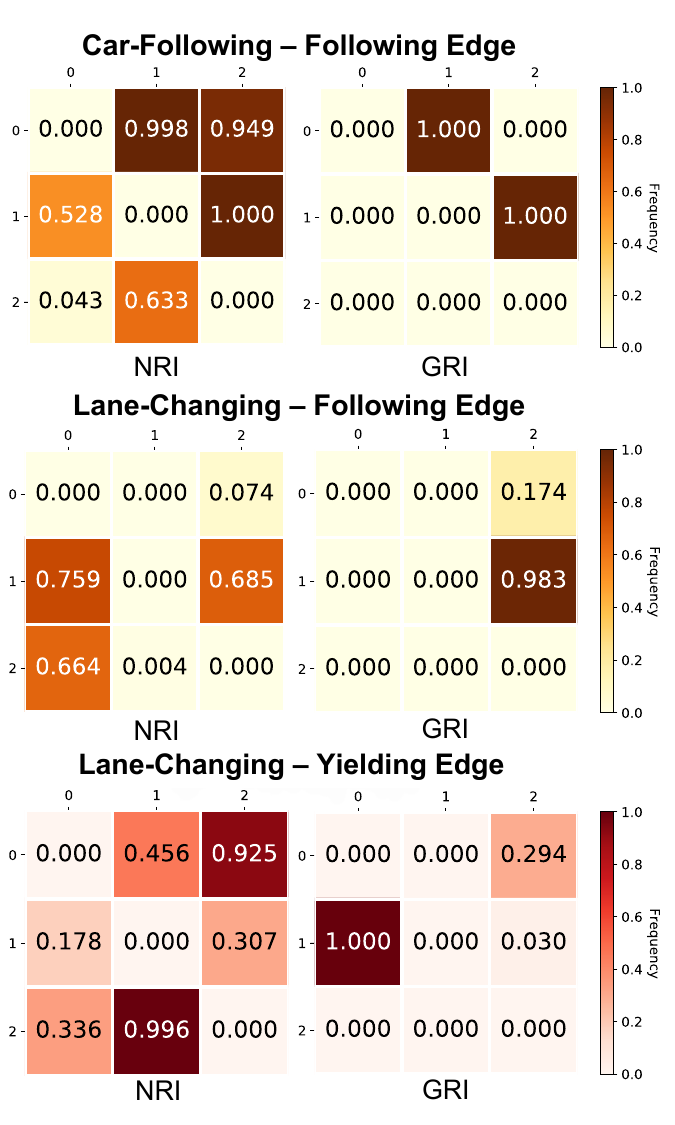} 
	\caption{The empirical distribution of estimated edge variables $\hat{z}$ over the test dataset in the naturalistic traffic scenarios. We summarize the results in multiple adjacency matrices corresponding to different edge types. In the adjacency matrix corresponding to the $k^\mathrm{th}$ type of interaction, the element $A_{i,j}$ indicates the relative frequency of $z_{j,i}=k$, where $z_{j,i}$ is the latent variable for the edge from node $j$ to node $i$. } \label{fig:ngsim_graph}
\end{figure}

We visualize the interaction graphs in Fig.~\ref{fig:ngsim_graph}. One interesting observation is that the graphs inferred by NRI have more edges in general. We want to emphasize that both models are trained under the same sparsity constraint. The results imply that we could guide the model to explore a clean and sparse representation of interactions by incorporating relevant domain knowledge, whereas the sparsity regularization itself is not sufficient to serve the purpose. Moreover, the NRI model assigns the same edge type to both edges between a pair of agents. It makes the graphs less interpretable because the vehicles ought to affect each other in different ways. On the other hand, even if different from the hypotheses, our GRI model tends to infer sparse graphs with directional edges.  

\textcolor{black}{For the supervised policy, it has the lowest reconstruction error in lane-changing. It implies that the human hypothesis is reasonable because it is capable to model the interactions among human drivers. For the car-following case, its reconstruction error is slightly higher than NRI. Since we cannot assure that our hypothesis is the ground-truth interaction graph underlying the interacting system\textemdash In fact, as we mentioned before, we never meant to treat it as the ground-truth\textemdash it is possible that the NRI model can find a latent space that can effectively model the interactions in an unsupervised manner. However, as shown in Fig.~\ref{fig:ngsim_graph}, it is difficult to interpret the graphs inferred by NRI. Considering the sparse and semantic nature of the hypothesis as well as the fact that the supervised policy's reconstruction error is on par with the NRI model, we think the chosen hypothesis is a valid one.}

\begin{table*}[t]
\centering
\caption{Performance Comparison on Naturalistic Traffic Dataset}
\label{table:ngsim}
\begin{threeparttable}
\centering
\begin{tabular}{|c|c|c|c|c|c|c|c|c|}
\hline
\multirow{2}{*}{Model} & \multicolumn{3}{c|}{Car Following ($\Delta t=0.2s, T=30$)} & \multicolumn{4}{c|}{Lane Changing ($\Delta t=0.2s, T=40$)} \\ \cline{2-8} 
& $\mathrm{RMSE}_x(\mathrm{m})$ & $\mathrm{RMSE}_v(\mathrm{m/s})$ &  $\mathrm{Accuracy}(\%)$ & $\mathrm{RMSE}_x(\mathrm{m})$ & $\mathrm{RMSE}_y(\mathrm{m})$ & $\mathrm{RMSE}_v(\mathrm{m/s})$ & $\mathrm{Accuracy}(\%)$   \\ \hline
GRI & $1.700\pm{1.005}$ & $0.721\pm{0.363}$ & $\mathbf{100.00\pm{0.00}}$ & $7.118\pm{3.647}$ & $0.764\pm{0.336}$ & ${4.320\pm2.392}$ & $\mathbf{98.55\pm{0.06}}$       \\ \hline
NRI & $\mathbf{1.436\pm{0.880}}$ & $\mathbf{0.650\pm{0.328}}$ & $64.09\pm{0.08}$ & $6.532\pm{3.822}$ & $0.330\pm{0.181}$ & $\mathbf{4.291\pm2.544}$ & $28.98\pm{0.08}$ \\ \hline
Supervised & ${1.482\pm{0.938}}$ & ${0.665\pm{0.344}}$ & - & $\mathbf{5.897\pm{3.651}}$ & $\mathbf{0.323\pm{0.223}}$ & $4.307\pm{2.435}$ & - \\ \hline
\end{tabular}
\begin{tablenotes}
	\item[1] The data is presented in form of $\text{mean}\pm{\text{std}}$.
\end{tablenotes}
\end{threeparttable}
\end{table*}

\subsection{\textcolor{black}{Semantic Meaning of Latent Space}} \label{sec:ood}
\textcolor{black}{The above experimental results show that our GRI model can recover the ground-truth interaction graphs in the synthetic scenarios with high accuracy, and infer interaction graphs that are consistent with human hypothesis on the NGSIM dataset. However, as we argue in Sec.~\ref{sec:introduction}, accurate interaction inference alone is not sufficient to show that the model learns a semantically meaningful latent space that is consistent with human domain knowledge. Given an edge, the policy decoder should also synthesize the corresponding semantic interactive behavior indicated by its edge type. It is difficult to verify whether the policy decoder is able to synthesize semantically meaningful interaction simply by monitoring the reconstruction error. Small reconstruction error on in-distribution data could be achieved by imitating demonstration without modeling the correct interaction~\cite{de2019causal, tang2021exploring}. To study the semantic meaning of latent space, we design a set of out-of-distribution tests \footnote{For clarification, the models used in this section are the same as those introduced in Sec.~\ref{sec:synthetic}. We merely designed additional out-of-distribution cases for testing.} by adding increasing perturbation to the initial states. We then enforce the same edge types as in the in-distribution case and run those different policy decoders to generate the trajectories. We are curious about whether the policy decoders can consistently synthesize the correct semantic interactive behavior under distribution shift. If so, we claim the latent space indeed possesses the semantic meaning that is consistent with human domain knowledge.}

\begin{figure}[t]
	\centering
	\includegraphics[width=2.4in]{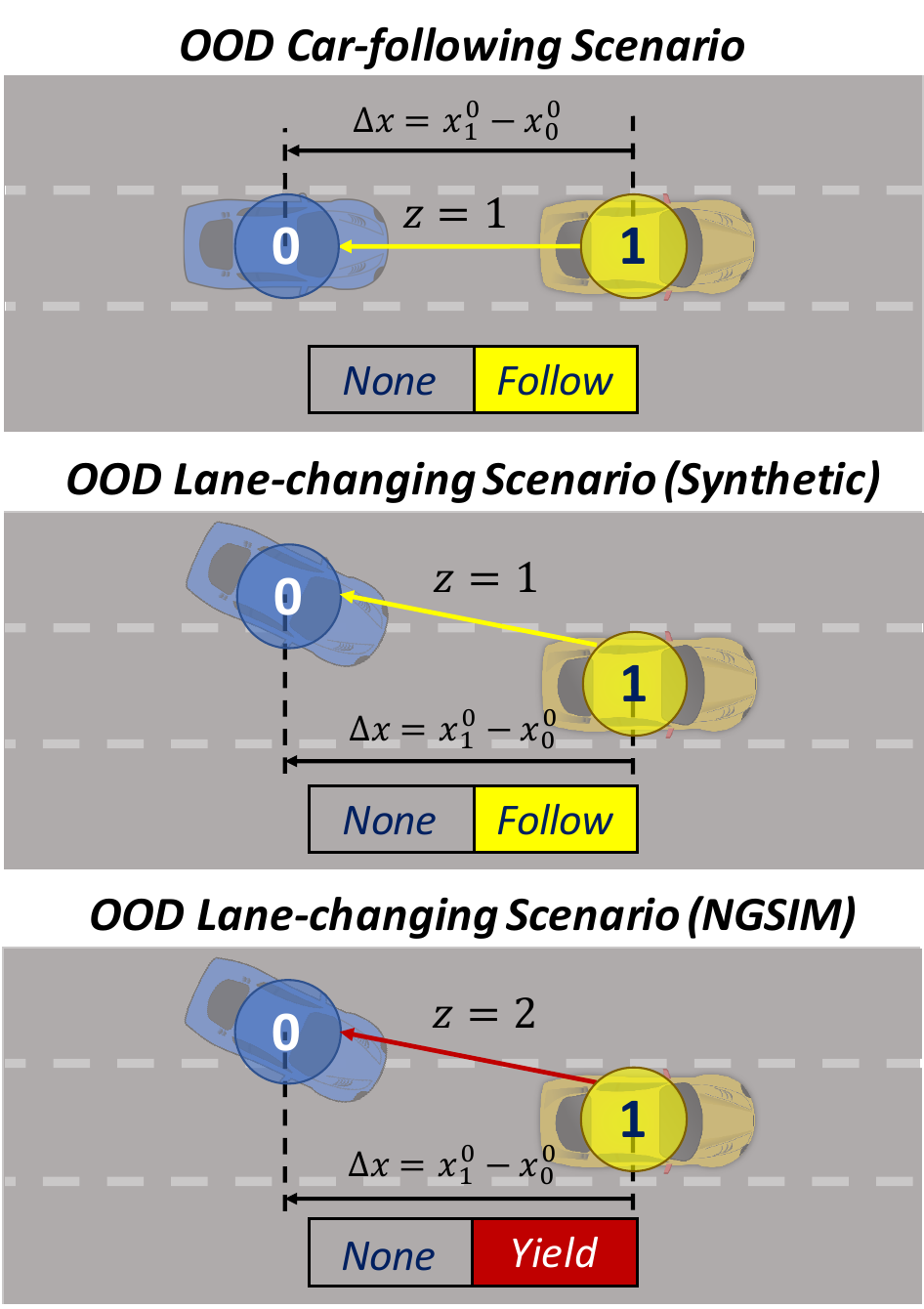} 
	\caption{Out-of-distribution scenarios. We removed one vehicle from the nominal scenes and shifted the initial longitudinal headway $\Delta x$ to unseen values.} \label{fig:ood_scene}
\end{figure}

\textcolor{black}{In the synthetic scenarios, we focus on the following relation. For both car-following and lane-changing scenes, we keep the two vehicles with the following relation, resulting in interaction graphs merely consisting of the following edges (Fig.~\ref{fig:ood_scene}). We introduce perturbation by decreasing the initial longitudinal headway to values unseen during the training stage.} The initial longitudinal headway is defined as $\Delta x=x^0_1 - x^0_0$, namely the longitudinal distance from Vehicle 1 to Vehicle 0 at the first time step. During the training stage, we sampled $\Delta x$ from uniform distributions: In car-following, $\Delta x\sim \mathrm{unif}(4, 8)$; In lane-changing, $\Delta x\sim \mathrm{unif}(8, 12)$. In the out-of-distribution experiments, we gradually decreased $\Delta x$ from the lower bound to some negative value, which means Vehicle 0 is placed in front of Vehicle 1. We are curious about if the models can generate trajectories meeting the characteristics of the car-following behavior in these unseen scenarios\textemdash scenarios with a different number of vehicles and distorted state distribution. To quantitatively evaluate if the synthesized behavior satisfies the requirement of car-following, we consider three metrics for evaluation:

\begin{itemize}
{\color{black}
    \item Success Rate: 
    \begin{align}
        \mathrm{Success Rate} &= \frac{1}{N}\sum_{i=1}^N\mathbf{1}(\Delta x^f_{i} \geqslant \delta_f), \label{eqn:success_rate} \\
        \textrm{where } \Delta x^f_{i} &= x^T_{1,i} - x^T_{0,i}, \nonumber
    \end{align}
    
    \item Collision Rate:
    \begin{align}
        \mathrm{Collision Rate} &= \frac{1}{N}\sum_{i=1}^N \mathbf{1}(d_{\min,i} \leqslant \delta_c),  \label{eqn:minimum_distance} \\
        \textrm{where } d_{\min,i} & = \min_t \sqrt{\left|x^t_{1,i} - x^t_{0,i}\right|^2 + \left|y^t_{1,i} - y^t_{0,i}\right|^2}, \nonumber
    \end{align}
}
    
    \item Lateral distance:
    \begin{equation}
        \Delta y = \left|y^T_1 - y^T_0\right| - \left|y^0_1 - y^0_0\right|. \label{eqn:lateral_distance}
    \end{equation}

\end{itemize}

We intend to quantify three typical characteristics of the following behavior with the metrics defined above: 1) staying behind the leading vehicle; 2) maintaining a substantial safe distance from the leading vehicle; 3) keeping in the same lane as the leading vehicle. \textcolor{black}{We consider the following vehicle's maneuver successful if the vehicle manages to keep a substantial positive final headway. And we consider two vehicles colliding if the minimum distance between them is smaller than a safety threshold.} Lastly, we expect the following behavior to attain a negative $\Delta y$, which means the following vehicle attempts to approach the leading vehicle's lane.  

All metrics were applied in the lane-changing scenario, but we only adopted $\mathrm{Success Rate}$ in the car-following scenario. Since we only model the longitudinal dynamics, $\Delta y$ is not applicable. For the same reason, if their initial positions are too close or the following vehicle is located ahead of the leading one initially, the following vehicle will inevitably crush into the leading vehicle, which results in $d_{\min}=0$. Therefore, we only care about the first characteristic.  

\begin{figure}[t]
	\centering
	\includegraphics[width=2.4in]{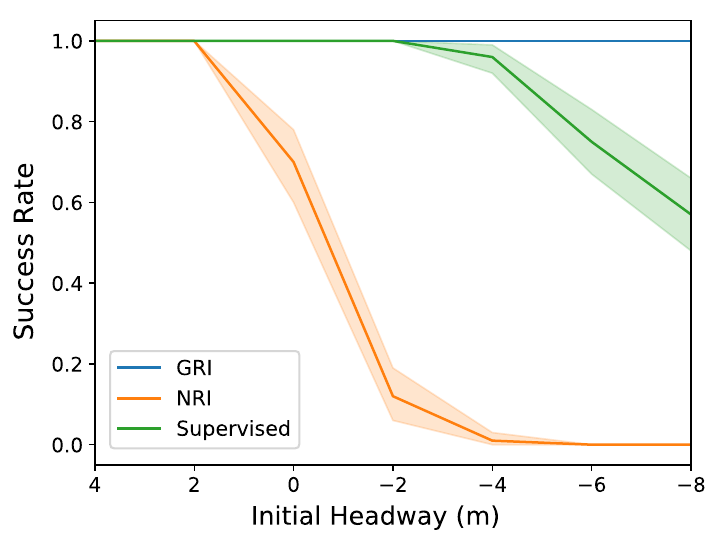} 
	\caption{\textcolor{black}{Results in out-of-distribution synthetic car-following scenario. We plot $\mathrm{Success Rate}$ versus $\Delta x$ with the error band denoting $95\%$ confidence interval of the indicator, $\mathbf{1}(\Delta x^f_i\geqslant \delta_f)$. We set $\delta_f=2\mathrm{m}$.}} \label{fig:ood_stats_cf}
\end{figure}

\begin{figure}[t]
	\centering
	\includegraphics[width=2.5in]{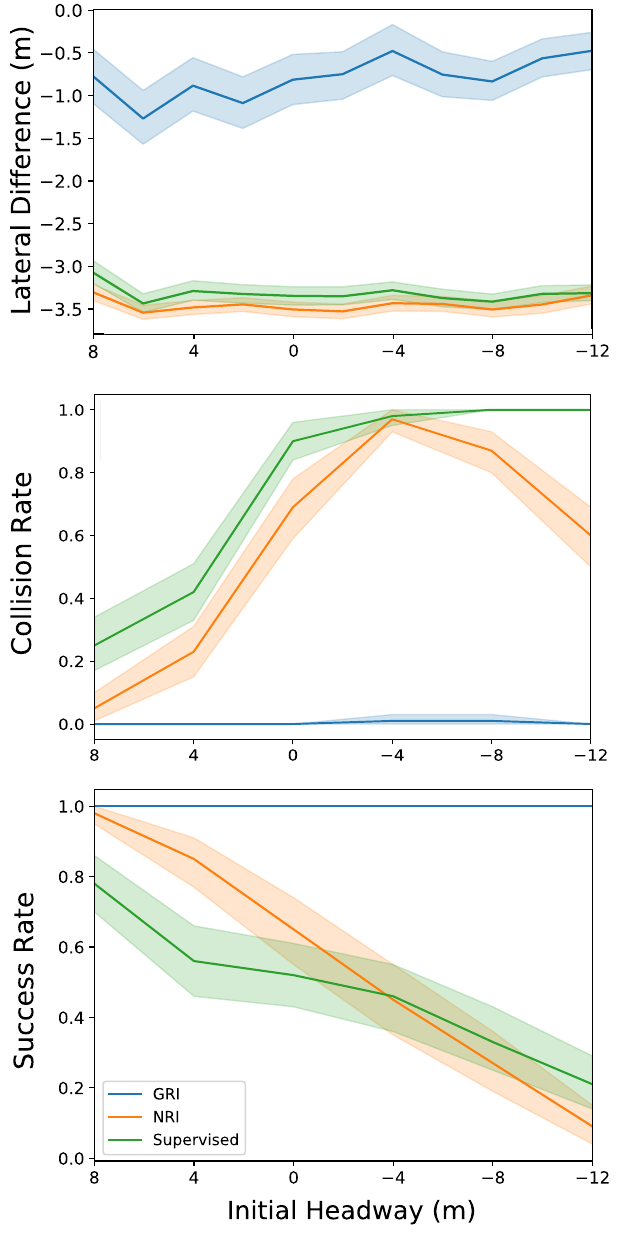} 
	\caption{\textcolor{black}{Results in out-of-distribution synthetic lane-changing scenario. We plot $\mathrm{Success Rate}$, $\mathrm{Collision Rate}$, and the mean value of $\Delta y$ versus $\Delta x$. The error bands denote $95\%$ confidence interval. For $\mathrm{Success Rate}$ and $\mathrm{Collision Rate}$, the error bands are of the indicator functions. We set $\delta_f=\delta_c=2\mathrm{m}$.}} \label{fig:ood_stats_lc}
\end{figure}

\begin{figure*}[t]
\centering
\includegraphics[width=6.2in]{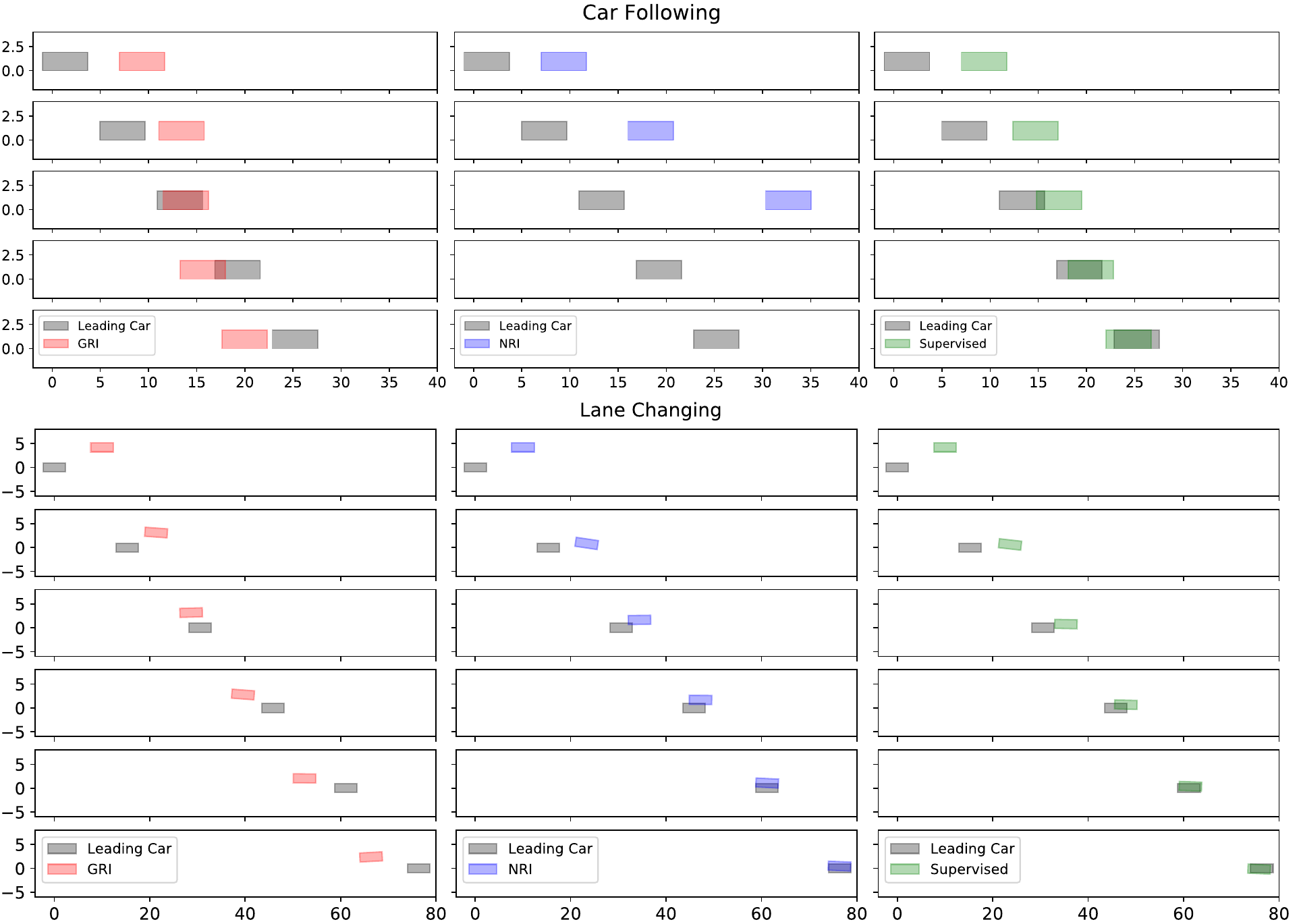} 
\caption{\textcolor{black}{Examples where the leading car is placed behind the following one at the initial timestep. The trajectories are visualized as sequences of rectangles. Each rectangle represents a vehicle at a specific time step. The vehicles are driving along the positive direction of the x-axis. The GRI policy still prompts the car-following behavior: It slows down the vehicle until the leading one surpasses it. Meanwhile, the NRI policy and the supervised one do not behave as $\mathcal{G}_\mathrm{interact}$ suggests.}} \label{fig:ood}
\end{figure*}

The results are summarized in Fig.~\ref{fig:ood_stats_cf} and Fig.~\ref{fig:ood_stats_lc}, where we plot the mean values of the evaluated metrics versus $\Delta x$. \textcolor{black}{In the car-following scenario, the NRI policy fails to slow down Vehicle 0 to follow Vehicle 1 when $\Delta x$ becomes negative. In contrast, the supervised policy and GRI policy maintain high success rates with negative $\Delta x$. However, the number of failure cases starts to increase for the supervised policy when $\Delta x$ becomes substantially negative, whereas the GRI policy maintains a perfect success rate over the tested range of perturbation.} We visualize a marginal example in Fig.\ref{fig:ood}, where both the NRI policy and the supervised one fail to maintain a positive final headway. 

\textcolor{black}{In the lane-changing scenario, the GRI policy maintains a consistent perfect success rate over all tested values of $\Delta x$. For the other two models, the success rates drastically decrease with decreasing $\Delta x$.} In terms of $\Delta y$, all models tend to reduce the lateral distance between the vehicles which is consistent with the second characteristic of the following behavior. However, we found that the GRI policy attains an average $\Delta y$ with a smaller magnitude and the magnitude decreases with decreasing $\Delta x$. It implies that the GRI policy changes its strategy when the initial position of Vehicle 0 is ahead of Vehicle 1. In order to keep a proper safe distance, Vehicle 0 does not change its lane until Vehicle 1 surpasses itself.
On the other hand, the lateral behavior is unchanged for the other two models. However, the vehicle cannot maintain a substantial safe distance if it changes its lane too early, \textcolor{black}{which is verified by the plot of collision rate versus $\Delta x$}. The difference in their strategies is further illustrated by the example visualized in Fig.~\ref{fig:ood}.  

\begin{figure}[t]
	\centering
	\includegraphics[width=2.4in]{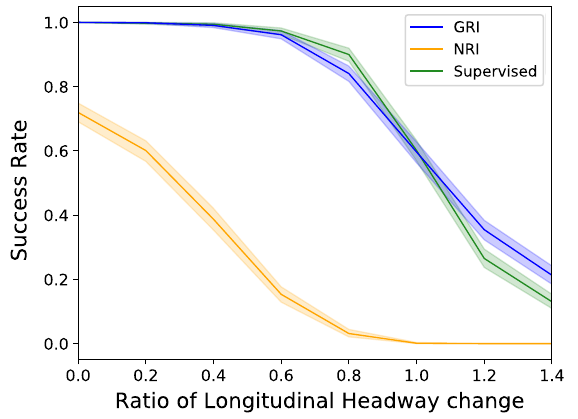} 
	\caption{\textcolor{black}{Results in out-of-distribution naturalistic traffic car-following scenario. We plot $\mathrm{Success Rate}$ versus $\Delta x$ with the error bands denoting $95\%$ confidence interval of the indicator, $\mathbf{1}(\Delta x^f_i\geqslant \delta_f)$. We set $\delta_f=2\mathrm{m}$.}} \label{fig:ood_stats_cf_ngsim}
\end{figure}

\begin{figure}[t]
	\centering
	\includegraphics[width=2.4in]{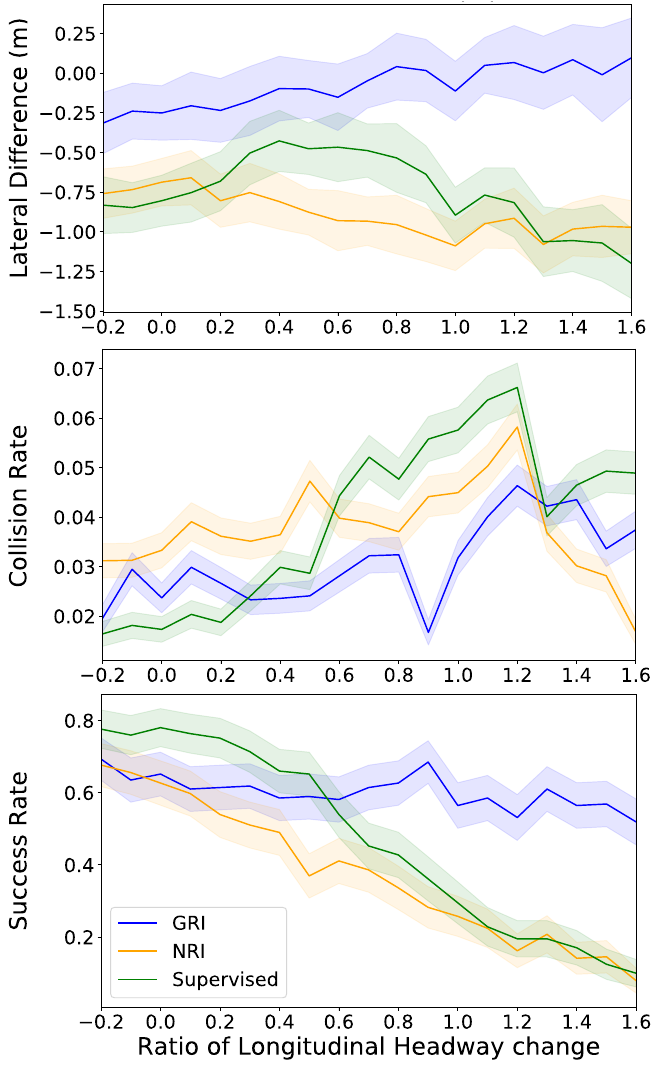} 
\caption{\textcolor{black}{Results in out-of-distribution naturalistic traffic lane-changing scenario. We plot $\mathrm{Success Rate}$, $\mathrm{Collision Rate}$, and the mean value of $\Delta y$ versus $\Delta x$. The error bands denote $95\%$ confidence interval. For $\mathrm{Success Rate}$ and $\mathrm{Collision Rate}$, the error bands are of the indicator functions. We set $\delta_f=\delta_c=2\mathrm{m}$.}} \label{fig:ood_stats_ngsim_lc}
\end{figure}

We repeat the experiment on the NGSIM datasets. Similar to the case with the synthetic dataset, we remove one vehicle from each scene, resulting in an interaction graph consisting of a single edge (Fig.~\ref{fig:ood_scene}). \textcolor{black}{It is worth noting that removing a vehicle from a scene alters the dynamic of the interacting system. It is not fair to expect the models to synthesize the same trajectories in the dataset. Therefore, we do not aim to compare the generated trajectories with the ones in the dataset in this out-of-distribution test. We just check whether the generated trajectories satisfy the desired characteristics of the corresponding interactive behaviors.}

In the lane-changing case, the remaining edge has the type of yielding. According to our definition of the yielding relation, we consider the same characteristics and adopt the same metrics defined in Eqn. (\ref{eqn:success_rate})-(\ref{eqn:lateral_distance}) for evaluation. Since we do not have control over the data generation procedure, we generate out-of-distribution test samples with different levels of discrepancy by controlling the ratio of longitudinal headway change. Given a sample from the original test dataset, we generate its corresponding out-of-distribution sample by shifting its initial longitudinal headway $\Delta x$ by a certain ratio, denoted by $\delta$, resulting in a new longitudinal headway $\Delta x'$:
\begin{equation*}
\Delta x' = (1 - \delta)\Delta x.
\end{equation*}
We evaluate the models on datasets generated with different values of $\delta$. We are particularly interested in the cases when $\delta\geqslant1$, which leads to a negative initial headway. We present the results in Fig.~\ref{fig:ood_stats_cf_ngsim} and $\ref{fig:ood_stats_ngsim_lc}$. The comparison is quite consistent with the synthetic scenarios. Compared to the other baselines, our GRI policy can synthesize trajectories that satisfy the desired semantic properties in a larger range of distribution shifts. 

The results suggest that even though the NRI model can accurately reconstruct the trajectories, the unsupervised latent space, and the corresponding policies do not capture the semantic meanings behind the interactions. In contrast, the GRI model learns a semantically meaningful latent space that is consistent with human domain knowledge. Another useful insight we draw from the experiment is that interaction labels are not sufficient to induce an explainable model with semantic latent space. Even though the supervised policy utilizes additional information on the ground-truth interaction graph, it fails to synthesize the following behavior in novel scenarios. Although the GRI model still has a considerable gap in reconstruction performance compared to the supervised baseline, it provides a promising and principled manner to incorporate domain knowledge into a learning-based autonomous driving system and induce an explainable model.

{\color{black}
\section{Discussion and Limitation}
\label{sec:discussion}
\subsection{Application of the Semantic Latent Space}
Enabling an explainable model is a crucial step toward trustworthy human interaction. However, it is still unclear how humans may benefit from improved explainability. We would like to have a brief discussion on the potential application of the semantic latent space introduced in GRI. When the autonomous vehicle encounters an unfamiliar situation (e.g., the out-of-distribution scenarios studied in Sec.~\ref{sec:ood}), a semantic latent space gives the safety drivers or passengers the privilege to review and override the inferred interaction graph if the model misunderstands the scenario. In contrast, humans can neither understand an interaction graph nor identify the correct edge types, if the learned interactive behaviors do not have explicit semantic meaning. Such kind of safety assurance could help build up safe and trustworthy cooperation between humans and autonomous vehicles.

However, it is impractical to keep the users monitoring the model output in real time. Instead, we can introduce an additional module to detect out-of-distribution scenes~\cite{filos2020can, SunL-RSS-21} and use the estimated epistemic uncertainty to decide when to query the end users. In~\cite{filos2020can}, the authors proposed an adaptive variant of their robust imitative planning algorithm, which incorporates such a unit. It is also a common practice for current autonomous driving companies to have human assistants for vehicles to query under abnormal situations.

\subsection{Limitation of the Learning Algorithm}
In our experiments, GRI always has higher reconstruction errors than NRI, especially on the synthetic dataset. One of the reasons is that reconstruction error is not directly optimized under the AIRL formulation. The objective function of NRI consists of a reconstruction loss, which essentially minimizes the Euclidean distance between the reconstructed trajectory and the ground-truth one. In other words, it directly minimizes the RMSE metrics used in our evaluation. In contrast, GRI adopts the objective function of AIRL, which also minimizes the distance between the trajectory pair. However, the distance is defined by the learned discriminator and is not necessarily equivalent to the Euclidean distance. In Appx.~\ref{app:airl_ablation}, we study two AIRL baseline models on the synthetic dataset. The results suggest that none of these AIRL-based approaches achieve the same reconstruction performance as NRI.  

Another reason is that the current learning algorithm is not quite stable, because of the adversarial training scheme we introduce when incorporating AIRL into the original NRI model. In typical AIRL settings, we may mitigate this problem by warm-starting the training with a policy network pre-trained through imitation learning or behavior cloning~\cite{finn2016guided, yu2019meta}. However, since we aim to learn a semantic latent space, warm-starting the training with a model with unsupervised latent space is not helpful. Alternatively, we may initialize the policy decoder with the supervised one. One issue is that it will change our current setting where human labels are not required. We will investigate this new setting in our future work, and develop a more stable training scheme to further optimize the performance of GRI. 

The structured reward functions also interfere with the stability of the learning procedure. Compared to the variant of GRI studied in Appx.~\ref{app:airl_ablation} with semantic reward functions removed, we found GRI is more sensitive to hyperparameters and prone to diverging if not carefully tuned. It is because although the structured reward functions are differentiable, it is not guaranteed that the reward functions can be stably optimized through gradient descent. In our future work, we will explore a more stable and robust learning scheme with those structured reward functions. 

\subsection{\textcolor{black}{Scaling to Large-Scale Traffic Scenarios}}
\textcolor{black}{In our experiments, we tested GRI in three-car interactive scenarios extracted from the NGSIM dataset. Here we discuss how we may extend GRI to model more sophisticated interactive traffic scenarios consisting of a large number of agents. In our current implementation, we assume a complete bi-directed scene graph. The dense graph makes the message-passing operations computationally expensive when the number of agents $N$ is large. The number of edges grows quadratically with the number of agents. Thus, the message-passing operations involve matrix multiplications over matrices with $O(N^3)$ non-zero elements. To alleviate the computation burden, one practical solution is to construct sparse scene graphs instead. We may leverage heuristics~\cite{sun2022m2i} or data-driven methods~\cite{tolstaya2021identifying, luo2023jfp} to \emph{coarsely} rule out non-interactive agent pairs from the scene graph. Note that we only need a coarse and conservative pre-processing of the scene graphs, and leave it for GRI to infer the remaining non-interaction edges. Thus, we can still avoid the expense and bias introduced by querying human labor to obtain finely-labeled traffic data. It is reasonable to expect that we can obtain a sufficiently sparse scene graph in most cases, since highly interactive scenes are rare in real-world traffic~\cite{tolstaya2021identifying}. To further enhance the modeling capacity, we may consider removing some inherent assumptions of GRI on the interactive systems. For instance, the current GRI framework assumes a static interaction graph over the time horizon. We will investigate how to incorporate dynamic graph modeling~\cite{li2020evolvegraph} into GRI in future work. Also, as discussed in Sec.~\ref{sec:formulation}, we plan to further generalize GRI to non-cooperative scenarios by removing the cooperative assumption.}

\section{Conclusion and Future Work}
\label{sec:conclusion}
In this work, we propose Grounded Relational Inference (GRI), which models an interactive system's underlying dynamics by inferring the agents' semantic relations. By incorporating structured reward functions, we ground the relational latent space into semantically meaningful behaviors defined with expert domain knowledge. We demonstrate that GRI can model interactive traffic scenarios under both simulation and real-world settings, and generate semantic interaction graphs explaining the vehicle's behavior by their interactions. Although we limit our experiments to the autonomous driving domain, the model itself is formulated without specifying the context. As long as proper domain knowledge is available, the proposed method can be extended naturally to other fields (e.g., human-robot interaction). \textcolor{black}{As discussed in Sec.~\ref{sec:discussion}, there are several technical gaps we need to bridge before extending the current framework to more complicated traffic scenarios and interactive systems in other fields. In future work, we will investigate these directions to improve the performance and scalability of GRI while maintaining the advantages of GRI as an explainable model}. 

\section{Appendix}
    \subsection{Graph Neural Network Model Details} \label{app:model}
    In terms of model structure, both the encoder and the policy decoder are built based on node-to-node message-passing~\cite{gilmer2017neural}, consisting of a node-to-edge message-passing and an edge-to-node message-passing:
    \begin{align}
        v\rightarrow e:\ \  \mathbf{h}^l_{i,j} & = f^l_e(\mathbf{h}^l_i, \mathbf{h}^l_j, \mathbf{x}_{i,j}), \label{eqn:gnn-1}\\
        e\rightarrow v:\  \mathbf{h}^{l+1}_{j} & = f^l_v(\sum\nolimits_{i\in\mathcal{N}_j}\mathbf{h}^l_{i,j}, \mathbf{x}_j), \label{eqn:gnn-2}
    \end{align}
    where $\mathbf{h}^l_i$ is the embedded hidden state of node $v_i$ in the $l^{\rm th}$ layer and $\mathbf{h}^l_{i,j}$ is the embedded hidden state of the edge $e_{i,j}$. The features $\mathbf{x}_i$ and $\mathbf{x}_{i,j}$ are assigned to the node $v_i$ and the edge $e_{i,j}$ respectively as inputs. $\mathcal{N}_j$ denotes the set of the indices of $v_i$'s neighbouring nodes connected by an incoming edge. The functions $f^l_e$ and $f^l_v$ are neural networks for edges and nodes respectively, shared across the graph within the $l^\mathrm{th}$ layer of node-to-node massage-passing.

    {\bf GNN Encoder.} The GNN encoder is essentially the same as in NRI. It models the posterior distribution as $q_\phi(\mathbf{z}\vert\boldsymbol{\tau})$ with the following operations:
    \begin{align*}
        \mathbf{h}^1_j & = f_{\mathrm{emb}}(\mathbf{x}_j), \\
        v\rightarrow e:\ \: \mathbf{h}^1_{i,j} & = f^1_e(\mathbf{h}^1_i, \mathbf{h}^1_j), \\
        e\rightarrow v:\ \ \ \mathbf{h}^{2}_{j} & = f^1_v\left(\sum\nolimits_{i\neq j}\mathbf{h}^1_{i,j}\right), \\
        v\rightarrow e:\ \: \mathbf{h}^2_{i,j} & = f^2_e(\mathbf{h}^2_i, \mathbf{h}^2_j), \\
        q_\phi(\mathbf{z}_{i,j}\vert{\boldsymbol{\tau}}) & = \mathrm{softmax}\left(\mathbf{h}^2_{i,j}\right),
    \end{align*}
    where $f_e^1, f_v^1$ and $f_e^2$ are fully-connected networks (MLP) and $f_{\mathrm{emb}}$ is a 1D convolutional networks (CNN) with attentive pooling.
    
    {\bf GNN Policy Decoder.} The policy operates over $\mathcal{G}_\mathrm{interact}$ and models the distribution $\boldsymbol{\pi}_\eta \left(\mathbf{a}^t\vert{\mathbf{x}^t, \mathbf{z}}\right)$, which can be factorized with $\pi_\eta\left(\mathbf{a}^t_j\vert{\mathbf{x}^t, \mathbf{z}}\right)$ as in Eqn. (\ref{eqn:facto}). We model $\pi_\eta$ as a Gaussian distribution with the mean value parameterized by the following GNN:
    \begin{align}
    v\rightarrow e:\ \ \ \ \Tilde{\mathbf{h}}^t_{i,j} & = \sum_{k=0}^{K}\mathbf{1}(z_{i,j}=k) \Tilde{f}^{k}_{e}(\mathbf{x}^t_i, \mathbf{x}^t_j), \label{eqn:policy1}\\
    e\rightarrow v:\ \ \ \ \ \ \mathbf{\mu}_j^t & = \Tilde{f}_v\left(\sum\nolimits_{i\neq j}{\Tilde{\mathbf{h}}^t_{i,j}}\right), \label{eqn:policy2} \\
    \pi_\eta\left(\mathbf{a}^t_j\vert{\mathbf{x}^t, \mathbf{z}}\right) & = \mathcal{N}(\boldsymbol{\mu}^t_j, \sigma^2\mathbf{I}) \label{eqn:policy3}.
    \end{align}
    
    Alternatively, the model capacity is improved by using a recurrent policy $\pi_\eta\left(\mathbf{a}^t_j\vert \mathbf{x}^t, \dots, \mathbf{x}^1, \mathbf{z}\right)$; Namely, the agents take actions according to the historical trajectories of the system. We follow the practice in~\cite{kipf2018neural} and add a GRU unit to obtain the following recurrent model:
    \begin{align}
    v\rightarrow e:\ \ \ \ \Tilde{\mathbf{h}}^t_{i,j} & = \sum_{k=0}^{K}\mathbf{1}(z_{i,j}=k) \Tilde{f}^{k}_{e}\left(\Tilde{\mathbf{h}}^{t}_i,\Tilde{\mathbf{h}}^{t}_j\right), \label{eqn:rnn1}\\
    e\rightarrow v:\ \ \ \Tilde{\mathbf{h}}^{t+1}_j & = \mathrm{GRU}\left(\sum\nolimits_{i\neq j} \Tilde{\mathbf{h}}^t_{i,j}, \mathbf{x}^t_j, \Tilde{\mathbf{h}}^t_{j} \right), \\
    \mathbf{\mu}_j^t & = f_\mathrm{out}\left(\Tilde{\mathbf{h}}^{t+1}_j\right), \\
    \pi_\eta\left(\mathbf{a}^t_j\vert \mathbf{x}^t, \dots, \mathbf{x}^1, \mathbf{z}\right) & = \mathcal{N}(\boldsymbol{\mu}^t_j, \sigma^2\mathbf{I}), \label{eqn:rnn4}
    \end{align}
    where $\Tilde{\mathbf{h}}^t_i$ is the recurrent hidden state encoding the historical information up to the time step $t-1$.
    
    {\color{black}
    \subsection{Reconstruction Visualization on Synthetic Dataset}\label{app:visual}
    
    \begin{figure*}[t]
	\centering
	\includegraphics[width=7.1in]{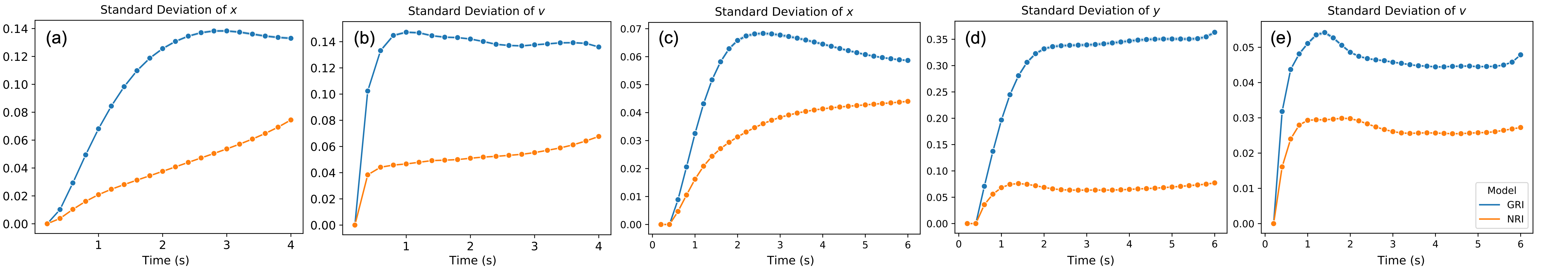} 
    \caption{\textcolor{black}{Average standard deviation of states along the time horizon. (a) and (b) show the standard deviation of $x$ and $v$ in the synthetic car-following scenario. (c)-(e) show the standard deviation of $x$, $y$, and $v$ in the synthetic lane-changing scenario.}}
    \label{fig:std_synthetic}
    \end{figure*}
    
    \begin{figure}[t]
	\centering
	\includegraphics[width=2.5in]{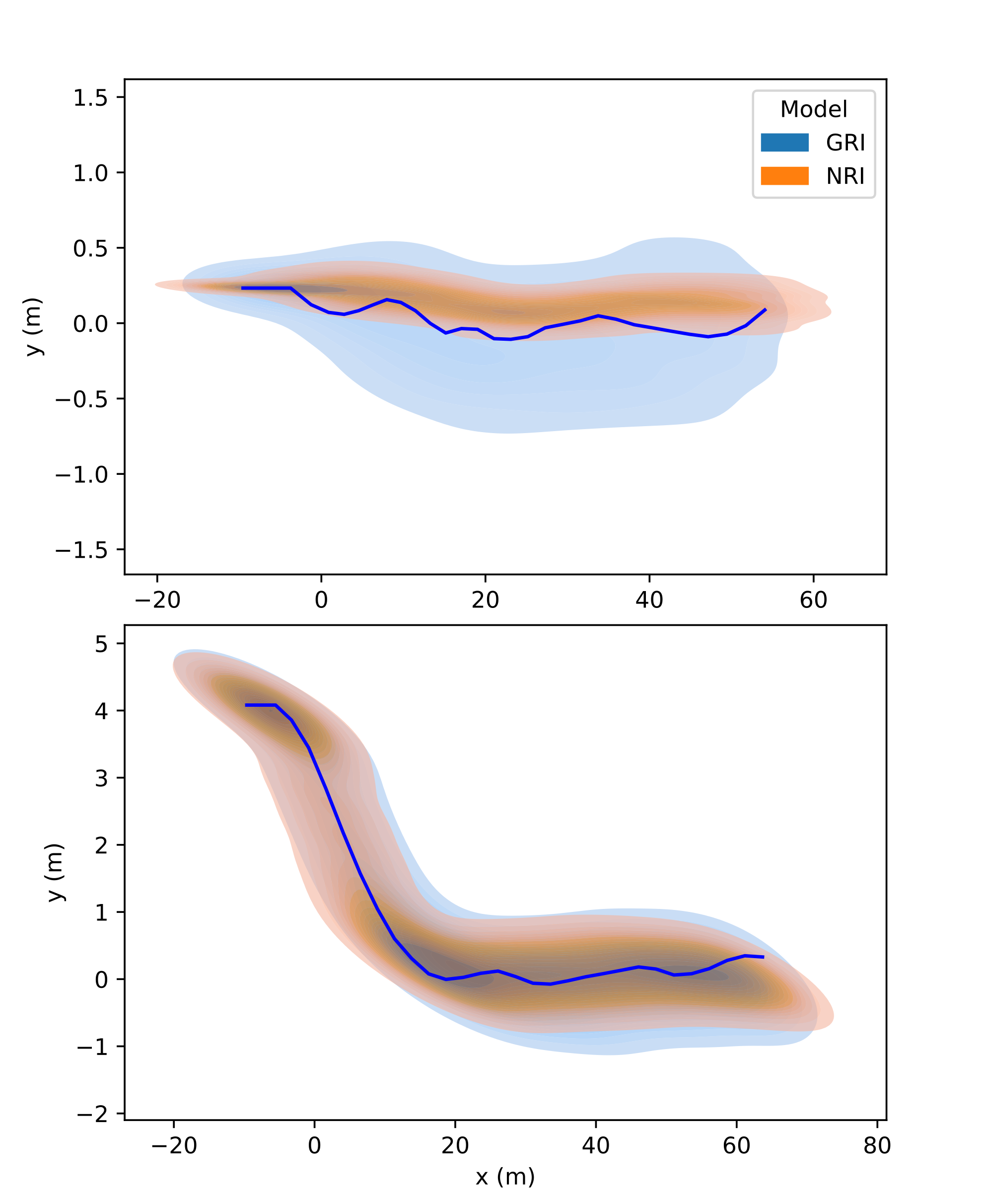} 
    \caption{\textcolor{black}{Visualization of the reconstructed trajectories in a lane-changing scene. (a) and (b) correspond to the trajectories of Car 1 and Car 0 respectively. We visualize the distributions of the reconstructed trajectories estimated using kernel density estimate. The ground-truth trajectories are in blue.}}
    \label{fig:visual_lc}
    \end{figure}
    
    In our experiments, we found that GRI has a significantly larger reconstruction error on the synthetic dataset than the NRI baseline. To better understand this performance gap in reconstruction, we looked into the reconstructed trajectories of both models. Instead of executing the mean value of the policy output as we did in our main experiments, we sampled the actions from the policy distribution to estimate the variance of reconstructed trajectories. In Fig.~\ref{fig:std_synthetic}, we plot the average standard deviation of reconstructed states along the time horizon. We observed that the policy decoder of GRI tends to have a larger variance. It partially explains the large RMSE values reported in Table~\ref{table:synthetic}: the metrics were computed with a single reconstructed trajectory. The policy distribution of GRI still has a larger bias than the one of NRI. We visualize the reconstructed trajectories of a lane-changing case in Fig.~\ref{fig:visual_lc}. While the GRI policy induces larger variance, the distribution of the reconstructed trajectories is sensible. 
    
    \subsection{AIRL Ablation Study} \label{app:airl_ablation}
    With the motivation of incorporating semantic meaning into the relational latent space, we developed GRI by introducing AIRL into relational inference and studied how the semantic reward functions may guide relational latent space learning. Meanwhile, it would be interesting to take a different perspective and study the effects of introducing relational inference and semantic reward functions into AIRL. In this section, we take the synthetic scenarios as examples and conduct an ablation study, where we compare GRI against two variants. 
    
    The first one is an AIRL variant, denoted by GRI-AIRL, which is obtained by removing relational inference and semantic reward functions from GRI. Concretely, both the policy and reward decoders operate on a fully-connected interaction graph with homogeneous edge types. And we simply use MLPs to model the reward functions in Eqn. (\ref{eqn:node_reward}) and (\ref{eqn:edge_reward}), instead of those semantic reward functions. The objective function then becomes Eqn. (\ref{eqn:opt-2}), but without neither the expectation over $\mathbf{z}$ nor the information bottleneck constraint. The second one is a variational AIRL variant, denoted by GRI-VAIRL, in which we introduce relational inference but do not use the semantic reward functions. In this case, the objective function is identical to the one in GRI, i.e., Eqn. (\ref{eqn:opt-2}). 
    
    \begin{figure}[t]
	\centering
	\includegraphics[width=2.5in]{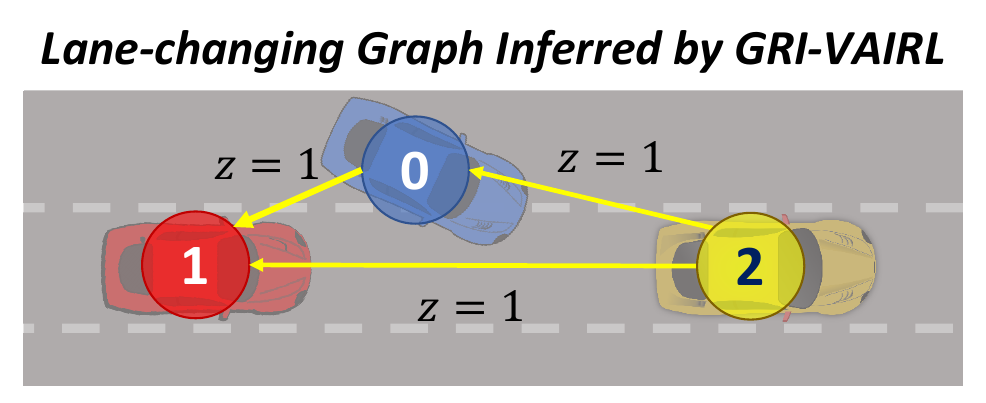} 
    \caption{\textcolor{black}{The interaction graph inferred by the GRI-VAIRL model in the synthetic lane-changing scenario.}}
    \label{fig:graph_vairl}
    \end{figure}
    
    The results are summarized in Table~\ref{table:ablation}. For the car-following scenario, the reconstruction performance is improved after introducing relational inference into AIRL. It is interesting that the GRI-VAIRL variant is able to recover the ground-truth interaction graph, even without the semantic reward functions. It makes sense because the car-following scenario only consists of a single non-trivial edge type. It is plausible for the model to distinguish non-interaction edges from others because a null reward is enforced for non-interaction edges. In some senses, we may still consider the reward function semantic\textemdash it incorporates the semantic meaning of non-interaction into the latent space. However, we cannot guarantee that GRI-VAIRL can distinguish between different non-trivial interactive behaviors, which is verified by the lane-changing case. Fig.~\ref{fig:graph_vairl} shows the inferred interaction graph. The model only adopts a single non-trivial edge type to describe all the interactive behaviors. Compared to the ground-truth graph, the inferred graph has an additional edge $z_{2,1}$ but ignores the edge $z_{1,0}$. Ignoring the edge $z_{1,0}$ limits the modeling capacity of the policy decoder, which could possibly explain why GRI-VAIRL has larger $\mathrm{RMSE}_x$ and $\mathrm{RMSE}_v$ than GRI-AIRL in the lane-changing case. 
    
    In summary, we could improve reconstruction performance by introducing relational inference into AIRL. Even if GRI-VAIRL has larger reconstruction errors in the lane-changing case due to the biased inferred graph, we still observe that GRI-VAIRL converges faster. The learning process becomes more stable and less sensitive to different hyperparameters. We think it is because the model may identify those agents that are not interacting with each other, preventing the reward decoder from fitting a reward function unifying both interactive and non-interactive behaviors. Meanwhile, it is still necessary to incorporate semantic reward functions to differentiate different interactive behaviors and induce a semantically meaningful interaction graph. However, semantic latent space comes at a cost of reconstruction performance. The structured reward functions limit the modeling capacity of the reward decoder. Also, although the structured reward functions are differentiable, it is not guaranteed that they can be well optimized through gradient descent. As a result, they may interfere with the stability of the learning procedure. 
    
    \begin{table*}[t]
    \centering
    \caption{\textcolor{black}{Ablation Study on Synthetic Dataset}}
    \label{table:ablation}
    \begin{threeparttable}
    \centering
    \begin{tabular}{|c|c|c|c|c|c|c|c|}
    \hline
    \multirow{2}{*}{Model} & \multicolumn{3}{c|}{Car Following ($\Delta t=0.2s$, $T=20$)} & \multicolumn{4}{c|}{Lane Changing ($\Delta t=0.2s$, $T=30$)} \\ \cline{2-8} 
    & $\mathrm{RMSE}_x(\mathrm{m})$ & $\mathrm{RMSE}_v(\mathrm{m/s})$ &  $\mathrm{Accuracy}(\%)$ & $\mathrm{RMSE}_x(\mathrm{m})$ & $\mathrm{RMSE}_y(\mathrm{m})$ & $\mathrm{RMSE}_v(\mathrm{m/s})$ & $\mathrm{Accuracy}(\%)$   \\ \hline
    GRI & $0.241\pm{0.125}$ & $0.174\pm{0.068}$ & $\mathbf{100.00\pm{0.00}}$ & $0.529\pm{0.230}$ & $0.207\pm{0.046}$ & $0.303\pm{0.128}$ & $\mathbf{99.95\pm{0.01}}$   \\ \hline
    GRI-VAIRL & $\mathbf{{0.120\pm{0.054}}}$ & $\mathbf{{0.116\pm{0.039}}}$ & $\mathbf{100.00\pm{0.00}}$ & ${0.377\pm{0.201}}$ & $\mathbf{{0.160\pm{0.038}}}$ & ${0.190\pm{0.058}}$ & $50.0\pm{0.00}$     \\ \hline
    GRI-AIRL & $0.138\pm{0.068}$ & $0.150\pm{0.043}$ & - & $\mathbf{0.304\pm{0.321}}$ & $0.198\pm{0.065}$ & $\mathbf{0.173\pm{0.101}}$ & -\\ \hline
    \end{tabular}
    \begin{tablenotes}
    	\item[1] The data is presented in form of $\text{mean}\pm{\text{std}}$.
    \end{tablenotes}
    \end{threeparttable}
    \end{table*}
    }

\bibliographystyle{IEEETran}
\bibliography{references}  

\end{document}